%% file: acl_latex.tex
\pdfoutput=1
\PassOptionsToPackage{prologue,dvipsnames}{xcolor}

\documentclass[11pt]{article}

\usepackage[final]{acl}

\usepackage{times}
\usepackage{latexsym}
\usepackage[most]{tcolorbox}

\usepackage[T1]{fontenc}
\usepackage[utf8]{inputenc}

\usepackage{microtype}

\usepackage{inconsolata}
\usepackage[textsize=scriptsize]{todonotes}

\usepackage{graphicx}
\usepackage{subcaption}
\usepackage{soul}
\usepackage{longtable}
\usepackage{comment}
\usepackage[dvipsnames]{xcolor}
\usepackage{multirow}

\makeatletter
\newcommand{\thickhline}{%
    \noalign {\ifnum 0=`}\fi \hrule height 1pt
    \futurelet \reserved@a \@xhline
}
\makeatother
\usepackage{ulem}
\usepackage{amsfonts}
\usepackage{mathtools}   % loads »amsmath«
\usepackage{makecell}
\usepackage{tabularx}

\usepackage{pifont}% http://ctan.org/pkg/pifont

\newcommand{\defeq}{\overset{\mathrm{def}}{=\joinrel=}}
\usepackage{arydshln}

\usepackage{bbm}
\usepackage{bm}
\usepackage{pgfplots}

\definecolor{myred}{RGB}{181, 18, 16}
\definecolor{myyellow}{RGB}{237, 144, 5}
\definecolor{myblue}{RGB}{66, 133, 244}

\newcommand\typewriter[1]{{\fontfamily{qcr}\selectfont #1}}
\title{CiteEval: Principle-Driven Citation Evaluation for  Source Attribution}

\author{
 \textbf{Yumo Xu$^{1}$} ~
 \textbf{Peng Qi$^{2\diamond*}$} ~
 \textbf{Jifan Chen$^{1*}$} ~
 \textbf{Kunlun Liu$^{1}$} ~
 \textbf{Rujun Han$^{3\diamond}$} \\
 \textbf{Lan Liu$^{1}$} ~
 \textbf{Bonan Min$^{1}$}~
 \textbf{Vittorio Castelli$^{1}$}~
 \textbf{Arshit Gupta$^{1}$}~
 \textbf{Zhiguo Wang$^{1}$}~
\\
\\
 \textsuperscript{1}AWS AI Labs~
 \textsuperscript{2}Orby.ai~
 \textsuperscript{3}Google~
\\
{\tt \{yumomxu,chenjf,kll,liuall,bonanmin,vittorca,arshig,zhiguow\}@amazon.com} \\
{\tt peng@orby.ai rujunh@google.com} \\
}

\begin{document}
\maketitle

\renewcommand{\thefootnote}{\fnsymbol{footnote}}
\footnotetext[1]{Equal contribution. $^{\diamond}$Work done at AWS AI Labs.}
\renewcommand{\thefootnote}{\arabic{footnote}}

\begin{abstract}
Citation quality is crucial in information-seeking systems, directly influencing trust and the effectiveness of information access. 
Current evaluation frameworks, both human and automatic, mainly rely on Natural Language Inference (NLI) to assess binary or ternary supportiveness from {cited} sources, which we argue is a suboptimal proxy for citation evaluation. In this work we introduce CiteEval, a citation evaluation framework driven by principles focusing on fine-grained citation assessment within a broad context, encompassing not only the cited sources but the full retrieval context, user query, and generated text.
Guided by the proposed framework, we construct CiteBench, a multi-domain benchmark with high-quality human annotations on citation quality.
To enable efficient evaluation, we further develop \textsc{CiteEval-Auto}, a suite of model-based metrics that exhibit strong correlation with human judgments. Experiments 
across diverse systems demonstrate \textsc{CiteEval-Auto}'s superior ability to capture the multifaceted nature of citations compared to existing metrics, offering a principled and scalable approach to evaluate model-generated citations.\footnote{Our code and datasets can be found at \url{https://github.com/amazon-science/CiteEval}}
\end{abstract}

\section{Introduction}
\label{sec:intro}
\input{fig_intro}
Information-seeking systems, such as retrieval-augmented generation (RAG; \citealt{lewis2020rag}) for question answering, play a vital role in how we access and understand knowledge. A key aspect of these systems is their ability to provide accurate  source attribution, typically in the form of citations \cite{gao-etal-2023-alce}. 
Accurate citations establish user trust and enable verification of generated content
\cite{liu-etal-2023-verifiability,malaviya-etal-2024-expertqa}.
Evaluating the quality of citations, however, remains a significant and under-addressed challenge.
Pioneered by Attributable to Identified Sources (AIS; \citealt{rashkin-etal-2023-ais}), existing work has largely focused on measuring the degree of \textit{supportiveness},
% from cited sources, 
using frameworks based on Natural Language Inference (NLI; \citealt{mnli2018williams}), for both human \cite{gao-etal-2023-alce,yue-etal-2023-attrscore} and automatic evaluation  \cite{zhang2024fine,fierro-etal-2024-learning}. 

In this work we critically examine the received wisdom and rethink whether NLI is truly the optimal lens through which to evaluate citation quality. 
As shown in Figure \ref{fig:intro}, NLI-based metrics determine the quality of citations solely based on the cited passages.
On the other hand, RAG systems often consume a wide range of contexts, 
and content in a model-generated response often come from paraphrasing the user query \cite{katsis2025mtrag}, fusing parametric knowledge from pre-training \cite{jiang-etal-2023-active}, or reasoning over preceding statements in the response \cite{trivedi-etal-2023-interleaving}. 
Not all can and should be attributable to the retrieved passages, and doing so, as a result of \textit{context insufficiency} in {evaluation},
leads to inaccurate estimation of citation quality.
Additionally, equating citation quality to binary or ternary {supportiveness} often falls short in capturing the nuances of citation utility \cite{malaviya-etal-2024-expertqa}. 
For instance, in Figure \ref{fig:intro}, 
the original citation that supports the \textit{``Concept of Inertia''} is from a sci-fi movie script, which can be supportive but less informative and credible compared to a physics book or Wikipedia page, as shown in the \textit{``Thinking''} and \textit{``Editing''} sections.

To address these limitations, we introduce CiteEval, a citation evaluation framework driven by a set of design principles, including:
\begin{enumerate}
    \item \textit{Evaluating citations against full retrieval sources} (Section \ref{subsec:principle_full_source}), 
    \item \textit{Evaluating citations beyond the retrieval context} (Section \ref{subsec:beyond_retrieval}), and
    \item \textit{Evaluating citations with fine-grained criteria and scenarios} (Section \ref{subsec:principle_fine_grained}).
\end{enumerate}
Particularly, CiteEval mitigates the context insufficiency in NLI-based approaches with Principles~1 and~2 jointly,
and moves beyond supportiveness and entailment in NLI by considering what constitutes good citations to {human} with Principle~3.
Guided by the proposed framework, we develop CiteBench, a high-quality citation evaluation benchmark to support research in this field.
CiteBench includes statement-level human judgments on RAG responses and citations, constructed with a multi-stage  annotation process comprising context attribution, critical editing, and citation rating.
To enable efficient assessment of citation quality, we further propose \mbox{\textsc{CiteEval-Auto}}, 
a suite of model-based metrics that demonstrate a strong correlation with human judgment.
We benchmark the citation quality of a wide array of existing systems, providing insights on the impact of critical RAG components, 
and the potential of CiteEval in citation improvement.

\section{Citation Evaluation Principles}
\label{sec:principles}

\subsection{Background}
\paragraph{Problem Formulation}
We first briefly introduce the task formulation for RAG. Given a corpus consisting of documents $\mathcal{D}$, a retriever fetches relevant sources $\mathcal{S}$ (e.g., fixed-length passages) for user query $Q$:
$\mathcal{S} = \text{Retriever}_\psi(\mathcal{D}, Q).$
A generator then reads the retrieved sources $\mathcal{S}$ to produce a response. 
Following \citet{gao-etal-2023-alce}, we employ an LLM-based generator which generates a text response $R$, \textit{jointly} with fine-grained citations $\mathcal C$: $R, \mathcal{C} = \text{Generator}_\chi(\mathcal{S}, Q)$.\footnote{Approaches to generate citations are many and varied (see Section \ref{sec:related_work} for details). We opt for joint generation considering its effectiveness and conceptual simplicity, but nothing prohibits this work from being applied to other approaches.}
As shown in Figure \ref{fig:intro}, when applicable, citations are provided at the end of each \textit{statement}: $\mathcal C=\{\mathcal C_i\}_{i=1}^{|R|}$ where $|R|$ denotes the number of statements in the response,
and statement-level citations $\mathcal{C}_i=\{c_{ij}\}$ is a subset of retrieved passages: $\mathcal C_i \subseteq \mathcal S$.

\paragraph{Attributable to Identified Sources (AIS)}
For the $i$th sentence in the response $R_i$, 
AIS \citep{rashkin-etal-2023-ais} evaluates its citations $\mathcal{C}_i$ as: 
\begin{gather}
r_i \defeq \text{NLI}_\phi(\text{concat }(\mathcal{C}_i), R_i) \label{eq:nli}
\end{gather}
where the rating $r_i \in \{0, 1\}$ denotes whether statement $R_i$ (i.e., the hypothesis) can be inferred from the concatenation of \textit{cited} sources $\mathcal{C}_i$ (i.e., the premise). 
As the rating approximates whether all required sources are cited, it has been adopted as citation \textit{recall} in AIS and its automatic version, Auto-AIS \cite{gao-etal-2023-rarr} based on an NLI model $\phi$. 
Auto-AIS further extends the notion to citation \textit{precision},
by applying $\phi$ to assess the relevance of each individual citation $c_{ij} \in \mathcal{C}_i$ \cite{liu-etal-2023-verifiability,gao-etal-2023-alce}.

\paragraph{CiteEval: A Principle-Driven Framework}
Departing from existing approaches based on NLI, such as Auto-AIS, 
we propose CiteEval which formulates the problem of citation evaluation as:
\begin{gather}
r_i \defeq f_\theta(\mathcal{C}_i; \mathcal{S}, R, Q).
\label{eq:citeeval_formulation}
\end{gather}
Particularly, $f_\theta$ directly estimates fine-grained ratings $r_i$ based on the full retrieval sources $\mathcal{S}$ and beyond (e.g., $R$ and $Q$). 
We then derive the response-level rating based on $\{r_i\}_{i=1}^{|R|}$.
We next state the principles mentioned in Section \ref{sec:intro} that we follow to drive the modeling of $f_\theta$. 

\subsection{{Evaluating Citations against Full Sources}}
\label{subsec:principle_full_source}
\input{tab_context_examples}

One underlying assumption of AIS is that \textit{self-evaluation} of $\mathcal{C}_i$ is sufficient, i.e., its rating is independent of the cite-worthiness of uncited sources $\mathcal{S}_i^-=\mathcal{S} \setminus \mathcal{C}_i$.
While source quality can be partially estimated based on its content, 
we note that its cite-worthiness should be determined in relation to other citation candidates.
For instance, citations will always receive the maximum AIS recall,
as long as they semantically entail the target response statement. 
This leads to over-estimation of the quality, when more reliable sources exist in the retrieval context and should be cited instead \cite{malaviya-etal-2024-expertqa}.
On the other hand, a citation will receive the lowest AIS precision score when it is partially supportive, even if there exists no better source that can fully support the statement. In this case, the score is underestimated as the citation can still be helpful in source verification, compared to the case where no citation is provided. 

To address these issues, we follow the principle that citation quality should be estimated against \textit{full} retrieval sources, i.e., all documents or passages retrieved for response and citation generation. This allows the model to leverage the quality of uncited sources for more accurate quality estimation for cited sources.
Additionally, full retrieval sources are also necessary in determining 
the boundaries with other contexts that should be considered in citation evaluation, which we will discuss next.

\subsection{Evaluating Citations Beyond Retrieval Context}
\label{subsec:beyond_retrieval}

Statements in the model response are often attributable to contexts beyond the retrieval context. For instance, in Figure~\ref{fig:intro}, ``\textit{You're asking about the significance of Newton's First Law of Motion}'' is simply a repetition of the user query and should not be attributed to the retrieved sources.
We consider three additional types of contexts beyond the retrieval sources: the \textit{user}, the \textit{response}, and the \textit{parametric knowledge}.

Table \ref{tab:context_examples} (upper block) presents a common case where responses start with a leading statement that repeats or paraphrases the query. 
These statements do not introduce new claims or facts beyond the context supplied by the \textit{user}, 
and are therefore not applicable for citation evaluation.
Autoregressive language models also condition token generation on their preceding statements in the \textit{response}, enabling {local reasoning} of varied types, from mathematical reasoning to planning.
Despite being topically related, statements shown in Table \ref{tab:context_examples} (middle block) are not directly 
conditioned on specific evidence in the retrieval context.  

The profound inherent knowledge of LLMs further allows \textit{parametric} facts to be incorporated in their outputs. 
These facts can be identified based on whether they are provided as part of the retrieval, user, or response context.
Formatting statements, such as transitional expressions in long-form answers \cite{fofo}, 
also fall into this category, considering the procedural knowledge that LLMs encapsulate.
When only parametric knowledge is involved in a statement, i.e., it is fully parametric, 
no citation should be possible given $\mathcal S$.\footnote{We discuss partially parametric cases in Appendix \ref{section:partially_parametric}.}

As contexts beyond retrieval are not citable, 
the above-mentioned statement types are \textit{not applicable} (\texttt{N/A}) for citation evaluation. 
In prior work, they are either implicitly penalized \cite{gao-etal-2023-rarr,gao-etal-2023-alce} or promoted \cite{zhang2024longcite} in existing citation evaluation frameworks. 
In this paper, we emphasize that all these contexts should be explicitly considered to enable accurate citation evaluation.

\subsection{Evaluating Citations with Fine-Grained Criteria and Scenarios}
\label{subsec:principle_fine_grained}
\paragraph{Criteria} 
Citations for the same statement can be inter-dependent.
To address this, NLI-based recall metrics 
evaluate citations for each statement as a whole, and 
project the combinatorial citation space into binary \cite{gao-etal-2023-alce} or ternary \cite{zhang2024longcite,yue-etal-2023-attrscore} supportiveness. 
This can potentially be too coarse when there are multiple citations and multiple evaluation aspects to consider.
Inspired by text generation where human evaluation is typically performed on a Likert scale \cite{xu2022latent,zheng2023judging}, 
we argue that citation evaluation can benefit from a more fine-grained evaluation schema, with rating guidelines incorporating relevant dimensions, such as citation redundancy and source credibility.

\input{tab_data_stats}

\paragraph{Scenarios}
Accessing the full contexts described in Section \ref{subsec:beyond_retrieval} helps evaluate the citation quality for all statements in the full response. 
However, we note that the full contexts may not always be observable to end users: commercial search engines such as Perplexity AI\footnote{\url{https://www.perplexity.ai}} and Microsoft Copilot\footnote{\url{https://copilot.microsoft.com}} only present \textit{cited} sources to users, rather than the full retrieval results.
In this scenario, users can only possibly focus on the statements that are cited, 
and highlighting the citation quality for these statements may better align with the user experience.
To this end, citation evaluation frameworks should cover the following two scenarios: 
1) \textbf{Full}, which evaluates all statements requiring citations, amongst which the uncited ones are penalized, and
2) \textbf{Cited}, where statements without citations, no matter whether citable, are treated as \texttt{N/A}.

\section{Building A Citation Benchmark with Principled Human Annotation}
\label{sec:citebench}

Driven by the outlined principles, we create CiteBench, a multi-domain citation evaluation dataset.
Particularly, CiteBench factorizes citation evaluation into three steps: context attribution, citation editing, and citation rating. 
Next we will introduce the human annotation  process, followed by the benchmark details. 

\subsection{Human Evaluation}
\label{subsec:human_eval}

\input{tab_human_correlation}

\paragraph{Process and Guidelines}

For context attribution, 
we provided human annotators a query, source passages retrieved by the query, and a model response consisting of one or more statements. 
Annotators were asked to first read all source passages and then attribute each statement to one of the major context types:
retrieval, user, response, or parametric. See Appendix \ref{appendix:context_attribuion_guidelines} for the complete guidelines. 

For statements attributed to the retrieval context, 
annotators were then asked to provide critical edits, to serve as rating evidences.
Particularly, we provided three deletion actions for different reasons: \typewriter{delete}{-misleading}, \typewriter{delete}{-substandard}, and \typewriter{delete}{-redundant}, 
as well as three addition actions: \typewriter{add}{-evidence}, \typewriter{add}-refinement, and \typewriter{add}-credibility.
Descriptions for these actions and detailed instructions were provided in Appendix \ref{appendix:edit_guidelines} as part of the annotation guidelines.
This allows edits to be performed with fine-grained reasons to impose varied impacts on citation ratings. For instance, deleting a misleading citation is likely to harm the rating more than a redundant one.
Each action operates on one target citation, which is either from existing citations (for \typewriter{delete} actions), or other retrieved sources (for \typewriter{add} actions). Citations that are not associated with an action are considered as \typewriter{keep}.

As the final step, we further asked annotators to rate the overall citation quality for each statement on a 1-5 Likert scale.
We present the rating guidelines in Appendix \ref{appendix:rating_guideline}, which
emphasize the fine-grained citation issues identified in the previous step, such as citation redundancy and missing evidence.

Critical editing and rating were skipped for statements attributed to contexts other than retrieval, which were labeled as not applicable (\texttt{N/A}).
We obtain response-level citation ratings via aggregating human ratings for statements that are applicable for citation evaluation with mean pooling.

\paragraph{Quality Control} 
Data annotation was performed by contracted data professionals with three blind passes. 
Across the three annotations for each sample, the context for each statement is determined using the majority vote, and the citation rating is determined using the average rating.
We have a dedicated team of data linguists to validate
the annotation quality. 
We performed three rounds of pilot annotation to fine-tune the taxonomy of context types and citation edits, addressing the ambiguities in the provided guidelines. 
The Inter-Annotator Agreement (IAA) of context attribution and citation rating are $0.980$ and $0.774$, respectively (The Krippendorff’s $\alpha$).

\subsection{Benchmark Dataset Construction}
\label{subsec:benchmark_creation}

\paragraph{Query Sampling and Passage Retrieval} 
We focus on Long-Form QA (LFQA) datasets for query sampling, as long answers for non-factoid queries can lead to more diverse citation behaviors for citation evaluation. Specifically, we include ASQA \citep{stelmakh-etal-2022-asqa}, ELI5 \cite{fan-etal-2019-eli5}, MS~MARCO \cite{msmarco}, and LFRQA \cite{ragqaarena}. Table \ref{tab:citebench_stats} summarizes the query distribution, featuring the coverage of multiple domains. 
Particularly, besides common knowledge corpora (e.g., Wikipedia and Bing), we cover queries from five emerging domains in LFRQA, including Science, Technology, Lifestyle, Recreation, and Writing.
We provide 10 fixed-size passages as the retrieval context for each query, obtained with various retrievers for each dataset to ensure contextual diversity. More details can be found in Appendix \ref{appendix:citebench}.
We uniformly sample in total 948 instances from MS~MARCO, ASQA and ELI5 as a development set and use the rest 3,000 instances for testing. 

\paragraph{Response and Citation Generation for Human Annotation} 
To perform human evaluation described in Section \ref{subsec:human_eval}, following \citet{gao-etal-2023-alce} we generate responses and citations in one pass with LLMs. 
We instruct citations to be generated in brackets at sentence end, which are then extracted with regex. The detailed prompt can be found in Appendix \ref{appendix:prompt_rag}. 
To control the annotation size, 
we randomly sample from ASQA, ELI5, and MS MARCO 100 queries each, and consider outputs from 4 models, including {GPT-4o} and {GPT-4o-mini}, {Llama-3-70b} and {Llama-3-8b} \cite{llama-3}, to balance proprietary and open-source models of varied sizes.
This yields a dataset of 1,200 instances for human annotation. 
We randomly sampled 200 instances for metric development and the rest 1,000 instances for meta-evaluation. 
We present the annotation statistics in Table \ref{tab:citebench_stats}.

\section{Model-Based Citation Evaluation}
\label{sec:mbe}
As human evaluation can be costly and time-consuming, we propose \textsc{CiteEval-Auto} to automate citation evaluation with model-based metrics. 

\subsection{Model-Based Context Attribution}
We instruct LLMs to attribute response statements to their contexts based on the context definition.
\citet{gao-etal-2023-alce} perform evaluation for each statement independently. For instance, the statement to be evaluated in Figure~\ref{fig:intro} ``\textit{It establishes the concept of inertia}''
is handled without co-reference resolution for \textit{It}.
To avoid the ambiguity, we provide all statements in one prompt, and instruct LLMs to iteratively attribute each statement to its context in one pass.
Citations are removed from responses to avoid introducing the bias of the retrieval context for cited statements (and vice versa).
Detailed prompt can be found in Appendix \ref{appendix:prompt_statement_classification}.

\subsection{Model-Based Citation Rating}
\input{fig_ablation}
For statements applicable for citation evaluation, we propose and discuss two rating approaches.

\paragraph{Iterative Chain of Edits (\textsc{IterCoE})} 
Given the citation rating guidelines, we instruct LLMs to follow a thinking step-by-step fashion by first reasoning about each statement against the given contexts.
We supply to LLMs \typewriter{delete} and \typewriter{add} actions introduced in Section \ref{subsec:human_eval}, allowing a sequence of edits to be generated for improving the citation quality.
LLMs then rate the target sentence on a 1-5 Likert scale, based on the generated edits and the rating guidelines.
Detailed prompts can be found in Appendix \ref{appendix:prompt_citation_editing}.
We normalize the estimated ratings to $[0, 1]$. 
We mask out the ratings for statements identified as \texttt{N/A} in context attribution, and aggregate the rest statement-level ratings into a response-level rating via mean pooling.

\paragraph{Edit Distance (\textsc{EditDist})}
Alternatively, we rate citations based on 
required edit actions and their estimated distances.
Particularly, different types of citation errors (e.g., a missing essential citation vs. a redundant one) should ideally impact the quality rating differently.
To this end, we learn an edit distance for each edit action type, based on how strongly the frequency of that action correlates with the overall human Likert score for a given statement. 
Specifically, let $\{a_k\}_{k=1}^K$ denote the edit actions defined in Section \ref{subsec:human_eval}, 
and $|\mathcal{A}^*_{i,k}|$ the number of occurrences of action $a_k$ in ground-truth actions $\mathcal{A}^*_i$.
We estimate the distance function $d(a_k)$ and overall rating $r_i$ via multiple linear regression $\text{min } \sum_i \text{MSE}(r_i, \hat{r}_i)$, where $\hat{r}_i$ denotes the human rating from the metric development set. 
The estimated rating $r_i$ is defined as:
\begin{gather}
    r_i = \sum_{k=1}^K 
    d(a_k) * \frac{|\mathcal{A}^*_{i,k}|}{|\mathcal{A}^*_{i}|} + b
\end{gather}
where $b$ is a bias term.
At test time, actions $\mathcal{A}_i$ are first generated using the same instructions as in \textsc{IterCoE} and then used for rating estimation.

\input{fig_estimated_edit_dist}

\subsection{Metric Evaluation Setup}
\label{subsec:metric_eval_setup}

We compare the performance of \textsc{CiteEval-Auto} against existing automatic metrics based on NLI:

\paragraph{\textsc{AutoAIS}~\cite{gao-etal-2023-alce}}
Unlike AutoAIS recall, \textsc{AutoAIS} Precision and F1 do not operate at the statement level. 
Apart from the original results, 
we tailor the framework to first produce statement-level precision and F1 scores, which are then averaged to response-level ratings (similar to our proposed approach). 

\paragraph{\textsc{AttrScore}~\cite{yue-etal-2023-attrscore}}
\textsc{AttrScore} evaluates each citation independently and classifies it as {attributable}, {extrapolatory}, or {contradictory}. 
We convert discrete categories into continuous ratings in two settings:
i) strict, which assigns a rating of 1 to \textit{attributable} and 0 to both \textit{extrapolatory} and {contradictory},
and 
ii) relaxed, which assumes \textit{extrapolatory} is to be relevant (but insufficient) which is assigned 0.5.

\paragraph{\textsc{LQAC}~\cite{zhang2024longcite}} 
\textsc{LQAC} (Long-Context QA with Citations) extends AutoAIS to include \textit{partial support} in precision and employs GPT-4o as the NLI model.
Similar to \textsc{AutoAIS}, we adapt it to first produce statement-level ratings and then response-level ratings.

\subsection{Metric Evaluation Results}
\input{tab_rag_results}

\paragraph{Human Correlation}
Table \ref{tab:results_metric_eval} shows the human correlation results in the \textbf{Full} scenario.\footnote{Results for the \textbf{Cited} scenario can be found in Appendix \ref{appendix:results_metric_eval_cited_only}. Details of the two scenarios are provided in Section \ref{subsec:principle_fine_grained}.}
% \rj{wrong table?}
\textsc{CiteEval-Auto} based on GPT-4o substantially outperforms state-of-the-art citation evaluators, at both the statement- and response-level.\footnote{We also experimented with other LLM backbones such as GPT-4-turbo and GPT-4o performs the best. Details are provided in Appendix \ref{sec:results_llms_for_citeeval}.}
We note that with {GPT-4o} as the backbone, \textsc{LQAC-Recall} achieves higher correlation compared to \textsc{AutoAIS-Recall}. 
Interestingly, \textsc{LQAC-Precision} does not yield better performance than its \textsc{AutoAIS} counterpart, although it is reported to achieve higher correlation with binary human labels on supportiveness. 

\paragraph{Ablation Study}
As shown in Table \ref{tab:context_attribution_results}, our proposed context attribution model yields $0.957$ F1 in predicting a statement's applicability for citation evaluation (see Appendix \ref{appendix:result_context_attribution} for a detailed performance breakdown).
To understand the effects of context attribution on final citation ratings, 
we further perform an ablation study and present the results in 
Figure~\ref{fig:citeeval_ablation} (left). 
We compare standalone citation rating models (e.g.,  \textsc{EditDist}) and their full \textsc{CiteEval-Auto} pipelines augmented with context attribution (e.g.,  \textsc{Ca}$+$\textsc{EditDist}). 
Removing context attribution causes substantial performance drops for both rating models we proposed.
Figure~\ref{fig:citeeval_ablation}  (right) further compares with the following approaches that directly rate citations without citation editing: \textsc{Vanilla} which directly rates all statements given the guidelines, 
\textsc{CoT} which performs a reasoning step before rating, as well as \textsc{IterCoT} which interleaves reasoning with citation rating. 
\textsc{IterCoE} outperforms these approaches by large margins,
showing the effectiveness of explicitly reasoning over the editing space and aligning model-generated edits with the rating guidelines.

We further show in Figure~\ref{fig:estimated_edit_dist} the estimated distance for each edit action. 
As we can see, \typewriter{add} actions lead to higher penalties compared to their \typewriter{delete} counterparts, demonstrating the necessity of identifying missing or better citation sources in citation evaluation.

\input{fig_effects_of_retrieval}

\section{Citation Benchmarking for RAG}

\paragraph{Models}
For a comprehensive automatic evaluation of LLMs on CiteBench, in addition to  GPT-4o and Llama-3 model families used in human evaluation (Section \ref{subsec:benchmark_creation}), we expand proprietary models to include GPT-4-turbo (2024-04-09) and GPT-3.5-turbo.
For public models, we further include two Mixtral models \cite{mixtral}: {Mixtral-8$\times$22B-Instruct} and {Mixtral-8$\times$7B-Instruct}, and two Qwen models \cite{qwen2025qwen25technicalreport}: Qwen2.5-72b and Qwen2.5-7b. 
We also benchmark LongCite-9B and LongCite-8B \cite{zhang2024longcite} which are fine-tuned for QA with citations.

\paragraph{Results}

We show in Table \ref{tab:rag_results} the benchmarking results. 
In the \textbf{Cited} scenario, GPT-4o is the top-performing model and Llama3-70b is on par with GPT-4o-mini.
On the other hand, 
in the \textbf{Full} scenario, Llama-3-70b outperforms GPT-4o and achieves the best performance.

To better understand the ranking differences in the two scenarios, we further examine the response and citation statistics from the benchmarked models, including the response length $|R|$ (i.e., average number of statements in a response) and missing ratio $M$ (i.e., average ratio of statements without citation). 
We found that GPT-4o tends to produce longer responses than Llama-3-70b.
% citations for some statements requiring citations.
To verify the impact of response length on citation behaviors, we measured the Pearson correlation between $|R|$ and $M$ which is $0.679$ ($p<.001$), indicating a strong positive correlation, i.e., longer responses are more likely to miss citations. 
This reveals the challenges in jointly generating long-text generation \textit{and} providing complete citations for all citable content. 
We provide more details on the correlation analysis in Appendix \ref{appendix:correlation_analysis}.

We also observed high missing citation ratios and substantially longer responses from fine-tuned LongCite models, leading to a large rating gap between the two evaluation scenarios.\footnote{In \citet{zhang2024longcite} fine-tuned models are shown to perform better than proprietary models, measured by \textsc{LQAC}.
We acknowledge that the conclusion cannot be drawn based on \textsc{CiteEval-Auto}.
One potential reason is \textsc{LQAC} assigns the highest rating to all statements that do not require citations. This leads to an over-estimation of citation quality for long responses wherein many \texttt{N/A} statements exist.}

\paragraph{Effects of Retrieval Quality}
\input{fig_iterative_improvement}
To examine how retrieval quality affects citation quality, we generate responses for ASQA and ELI5 with re-ranked retrieval contexts which have substantially higher recall of relevant passages.
For MS MARCO, as the original corpus is not accessible for reranking, we maximize the retrieval precision via filtering the retrieval context and keeping only passages annotated as relevant by human.
Figure \ref{fig:effects_of_retrieval} shows that using re-ranked contexts with higher retrieval recall often leads to higher citation quality in both evaluation scenarios.
On the other hand, citation quality does not benefit from better retrieval precision via filtering on MS MARCO.

\paragraph{Citation Improvement with Edit Actions}
Towards exploring the potential of \mbox{CiteEval} in citation improvement, we iteratively generate and execute edit actions and examine the rating dynamics.
Specifically, for a set of citations $\mathcal{C}^{(t)}$, we leverage \textsc{CiteEval-Auto} to jointly generate edit actions $\mathcal{A}^{(t)}$ and citation ratings $r^{(t)}$. 
We execute the actions against $\mathcal{C}^{(t)}$ to generate a set of new citations $\mathcal{A}^{(t)}: \mathcal{C}^{(t)} \mapsto \mathcal{C}^{(t+1)}$.
We repeat the process $T$ times, and report the citation rating $r^{(t)}$ at each iteration in Figure \ref{fig:iterative_improvement}.
We observe that CiteEval consistently improves citation quality across models, where larger models such as GPT-4o reach the performance peak quicker than smaller models. 
Regardless of the initial performance and model size, models within the same family converge to similar performance after a sufficient number of iterations. 
This opens up opportunities to improve small LLMs' source attribution performance with the executable critique from \textsc{CiteEval-Auto} and inference-time scaling \cite{snell2024scalingllmtesttimecompute}. 

\section{Related Work}
\label{sec:related_work}
The increasing demand for the deployment of LLMs in information-seeking systems has spurred efforts in source attribution, with external evidences presented as URLs \cite{muller-etal-2023-evaluating}, snippets \cite{gao-etal-2023-rarr}, quotes \cite{menick2022teachinglanguagemodelssupport}, or retrieved sources \cite{gao-etal-2023-alce}. 
Regardless of the evidence presentation, NLI is commonly adopted to judge the attribution quality \cite{liu-etal-2023-verifiability,zhang2024longcite}. 
\citet{yue-etal-2023-attrscore} introduced two evaluation sets for source attribution, derived from existing QA data and AI search engines. The evaluation sets assume citations are provided independently and evaluate only the leading sentence of a response. In this work, we focus on improving the evaluation of \textit{in-line} citations to retrieved sources, with a high-quality benchmark consisting of fine-grained human annotations for full responses.

\section{Conclusion}
We proposed CiteEval, a principle-driven framework centered on fine-grained citation ratings within comprehensive evaluation contexts.
Based on CiteEval, we constructed a high-quality citation evaluation dataset CiteBench, and proposed \textsc{CiteEval-Auto}, an automated metric for scalable evaluation. 
Experiments across diverse RAG systems highlight \textsc{CiteEval-Auto}'s enhanced ability in evaluating and improving citation quality.

Directions for future work are many and varied. One research challenge is to develop distillation techniques to approximate CiteEval judgments with smaller LMs.
We would also like to extend the proposed framework to RAG reward modeling and post-training, and enhance the trustworthiness of AI responses through effective knowledge grounding and attribution.

\section{Limitations}

Context attribution in this work focuses on typical contexts in RAG and can be expanded to cover more diverse use cases. For example, user's demographic information such as age and location is often used for more personalized responses \cite{zhang2024personalization}, which can also be considered as part of the user context in addition to queries. Also, context attribution is introduced as a sub-task for citation evaluation in this work. The task can be further applied to citation \textit{generation}, to move beyond attribution to the retrieval context and enable information verification from broader contexts.
For instance, citations to contexts beyond retrieval could be provided through special tokens denoting the context (such as \texttt{[P]} for parametric knowledge), or in the form of natural language. 

While our work establishes a strong correlation between \textsc{CiteEval-Auto} and fine-grained human judgments, a comprehensive evaluation of its real-world, downstream impact is a crucial next step. Such \textit{extrinsic} evaluation 
requires user studies or task-based evaluations that reliably measure constructs such as user trust and verification experience across downstream applications.
We believe that CiteEval provides a necessary foundation by offering a more reliable and principled \textit{intrinsic} evaluation of citation quality, paving the way for future studies into its \textit{extrinsic} impact.
 
Additionally, in this work we treat sentences as statements, a notion that can be extended to cover finer-grained text chunks.\footnote{Chunk-level citations are now supported by Anthropic API, subsequent to this work's completion: \url{https://www.anthropic.com/news/introducing-citations-api}} 
Towards evaluating citations in a more end-to-end setting, the retrieval step in RAG can be further incorporated in the proposed framework, potentially with ground-truth annotations on the relevance of retrieval sources.

\bibliography{custom}

% \clearpage
\appendix

\section{Human Annotation Instructions}
\label{appendix:annotation_guidelines}

\subsection{General Guidelines}
Your task is to evaluate the quality of citations generated by language models based on a given query, a set of retrieved passages relevant to that query, and the model's generated text containing the citations. Below, we define what is meant by the query, retrieved passages, and citations:

\begin{description}
    \item[Query] Usually an information-seeking question like ``What is the significance of Newton's First Law of Motion?''.
    \item[Retrieved Passages] For each query, you will be given a few (maximum 10) relevant passages from the indexed corpus.
    \item[Model Generation] For each query, the language model will generate a response based on the retrieved passages.
    \item[Citation] The language model will generate fine-grained citations at the sentence end in responses. Citations are represented as bracketed numbers, such as \texttt{[1][2][3]}. For each sentence in a response, its citations link to a subset of retrieved passages, if there are any.
\end{description} 
We break model answers into sentences for fine-grained evaluation. For each sentence in the answer, you will be asked to perform maximum three steps: context attribution, citation editing, and citation rating. Specifically, for each answer sentence: 

\begin{enumerate}
    \item Context Attribution: You will be asked to classify the sentence to one of four context types.
    \item Citation Editing: Depending on the context you attribute the sentence to, you may be asked to edit the provided citations to make them better.
    \item Citation Rating: You will be asked to rate the quality of the provided citations, based on the edits you may have performed in Step 2.
\end{enumerate}
Your annotation will be used in a project aiming to develop metrics that better reflect citation quality.

\subsection{Context Attribution}
\label{appendix:context_attribuion_guidelines}

First read all passages and pay special attention to evidence for the given question and each answer sentence. Then classify each answer sentence into one of the context types shown below. 

\begin{description}
\item[Query] Sentences that iterate or rephrase the user query without making new claims or involving new facts.\newline

\item[Retrieval] Sentences fully or partially supported by the retrieval context.

\item[Response] Sentences solely derived from preceding sentences within the response itself, not relying on the query context, the retrieval context, or the succeeding sentences in the response. Examples include sentences that perform mathematical and logical reasoning over preceding response sentences. 

\item[Model] Sentences solely based on the inherent knowledge of the language model that generated the response. Knowledge is only inherent when it can NOT be found in, or reasonably inferred from, the query context, the retrieval context, or the response context. Examples include unsupported facts, and transitional expressions/summarization without any substantial claims.
\end{description}

\subsection{Citation Editing}
\label{appendix:edit_guidelines}
\input{tab_edit_description}

Perform a few edits improve the quality of the citations using your best judgment. Each edit operates on one citation, and can be \typewriter{delete} or 
\typewriter{add} with specific reasons (see Table \ref{tab:edits}). 
You can add an edit with the $+$ button.
Edit citations based on Editing Guidelines:

\begin{itemize}
    \item If you think the citation is perfect, you don’t need to do anything.
    \item Add 0 as the citation ID for facts that can NOT be found in, or reasonably inferred from, the user query, the retrieved passages, or the model response. This attributes the unsupported facts to inherent knowledge of the language model that generated the response.
    \item You should aim to achieve citations of the highest standard with minimal editing. After editing, all major claims in the statement should be cited.
    \item After editing, the citations should cite sources that are mostly helpful, when there are multiple related sources. The final citations for each sentence typically contain at most 3 citations, but there can be exceptions (e.g., if more than 3 citations all include direct and complementary supporting evidence, they should all be included).
\end{itemize}

\subsection{Citation Rating}
\label{appendix:rating_guideline}
Review your edits if there is any. Based on the rating guidelines below, rate the quality of the original citations (NOT the citations after editing) from 1-5:

\begin{description}
    \item[5 (Excellent)] The sentence is fully supported by all relevant and accurate citations. There are no unnecessary, misleading, or missing citations. The citations (if present) enhance the credibility and informativeness of the sentence.
    \item[4 (Good)] The sentence is mostly supported by accurate and relevant citations. One potentially relevant citation may be missing, or a slightly unnecessary citation may be present, but these do not significantly detract from the overall quality of the sentence.
    \item[3 (Fair)] The sentence has some issues with citations. There might be one or few noticeable missing citation that somewhat weaken the sentence's support, or there might be several unnecessary or inaccurate citations that detract from the sentence's clarity or conciseness. Overall, the sentence's accuracy and credibility are somewhat compromised.
    \item[2 (Poor)] The sentence has significant problems with citations. There might be multiple missing citations that leave that leave central claims unsupported, or there might be multiple unnecessary or inaccurate citations that significantly undermine the sentence's accuracy and credibility.
    \item[1 (Unacceptable)] The sentence is completely unsupported by citations or is supported entirely by inaccurate, irrelevant, or misleading citations. The sentence is rendered misleading and unreliable.\newline
\end{description}

\section{Discussion: Partially-Parametric Statements}
\label{section:partially_parametric}
% \paragraph{Special Case: Partially Parametric}
One typical scenario is the fusion of parametric knowledge and retrieval contexts in one statement. 
Consider a {partially}-parametric statement: \textit{A hub simply repeats everything it hears, whereas a switch is a more intelligent device that can identify and direct traffic \texttt{[1]}}. 
The statement can be decomposed into the following two claims: 
\begin{enumerate}
    \item A hub simply repeats everything it hears, and
    \item A switch is a more intelligent device that can identify and direct traffic. 
\end{enumerate}
In this case, \texttt{[1]} is the best possible citation from the retrieved sources and supports Claim 1. 
On the other hand, no retrieved source supports Claim 2 (and neither do the user and response contexts), attributing Claim 2 to the parametric context.
% although one passage mentions that switches are different from hubs. 
% is from the parametric context.
% One option is to assign the highest rating to the best possible citations, considering its contribution to {verifiability} \cite{liu-etal-2023-verifiability}. 
We argue that the upper bound for this statement's citation rating is always lower than the highest rating defined in the rating schema. 
Even with the best possible citations, the user will not be able fully verify the statement (i.e., the statement remains partially supported), and will likely require extra efforts for a complete fact checking \cite{liu-etal-2023-verifiability}.
Also, providing any citations for partially supported statements may mislead users into trusting the whole statement, as citations naturally build credibility especially when users do not always check them \cite{lipson2011cite}. 
Leaving this type of statements uncited does not resolve this issue, as it renders the statement appear to be completely unsupported, which is neither optimal for its verifiability nor credibility.
CiteEval treats 0 as a special citation ID for parametric facts, and annotators or models can choose to add 0 as missing evidence when appropriate, and take it into account in the final rating (see Appendix \ref{appendix:edit_guidelines}).

\section{CiteBench Details}
\label{appendix:citebench}

\subsection{Retrieval Settings}
For LFRQA \cite{ragqaarena}, we follow the same retrieval setting which splits passages into text chunks with 100 consecutive words, and use the top 10 retrieved passages retrieved by {ColBERTv2} \cite{santhanam-etal-2022-colbertv2}. 
We randomly sample 1,000 instances from MS MARCO \cite{msmarco} which uses 10 passages from Bing logs and consists of 80\% answerable queries and 20\% unanswerable queries. 
For ASQA \citep{stelmakh-etal-2022-asqa} and ELI5 \cite{fan-etal-2019-eli5}, 
we follow \citet{gao-etal-2023-alce} and use the same subset, with top 10 passages retrieved by DPR \cite{karpukhin-etal-2020-dense} and BM25, respectively.

\subsection{Responses Post-processing}
\input{fig_response_token_dist}
We remove the thinking section in model responses via matching the start token \texttt{<thinking>} and end token \texttt{</thinking>}. 
Figure \ref{fig:token_dist} shows the response length distribution.
We split responses into statements with NLTK sentence tokenizer (version: 3.8.1).\footnote{\url{https://www.nltk.org}}
We extract citations from each statement with regex \verb/\[(\d+)\]/, and keep only citations in the indices of retrieved passages $[1, 10]$. 

\subsection{Dataset License}
We provide license information for the datasets used in this work to construct CiteBench as follows:

\begin{itemize}
    \item ASQA \cite{stelmakh-etal-2022-asqa}: Apache 2.0 License, \url{https://github.com/google-research/language/blob/master/LICENSE}
    \item LFRQA \cite{ragqaarena}: Apache 2.0 License, \url{https://github.com/awslabs/rag-qa-arena?tab=Apache-2.0-1-ov-file#readme}
    \item ELI5 \cite{fan-etal-2019-eli5}: BSD License, \url{https://github.com/facebookresearch/ELI5?tab=License-1-ov-file#readme}
    \item MS MARCO \cite{msmarco}: CC BY 4.0 License, \url{https://microsoft.github.io/msmarco/LICENSE}
\end{itemize}

\subsection{Annotation Details}
Human annotation was performed by contracted data professionals through Summa Lingual.\footnote{\url{https://summalinguae.com/}}
The rate is \$$31.5$ per annotation task, which includes three blind passes for one sample, and the total cost for the official annotation batch of $1,200$ samples is \$$37,800$.
The annotation was audited and finalized by a team of full-time data linguists and scientists based in the United States.

\section{Further Analysis}

\subsection{Effects of LLM Backbones for Citation Evaluation}
\label{sec:results_llms_for_citeeval}
\input{tab_results_citeeval_llms}

We show the performance of different LLMs for \textsc{CiteEval-Auto} in Table \ref{tab:citeeval_llms}.

\subsection{Results on Context Attribution}
\label{appendix:result_context_attribution}
\input{tab_human_correlation_star}

Table \ref{tab:context_attribution_results} shows the precision, recall, and F1 for context attribution in predicting whether a statement is applicable for citation evaluation, where the retrieval context maps to  \textit{Applicable}, while the user, response and parametric contexts are aggregated into the same \textit{Not Applicable} class as they are treated with an identical evaluation strategy in this work.

Figure \ref{fig:conf_mat} further provides a breakdown of the model predictions. As can be seen, one of the major error categories for context attribution is between the parametric and retrieval contexts, which is not surprising as faithfulness evaluation and hallucination detection are challenging tasks yet to be resolved \cite{zhang-etal-2024-fine}.

\subsection{Benchmarking Result Analysis}
\label{appendix:correlation_analysis}
\input{fig_correlation}
We show the correlation between the response length, missing-citation ratio, and citation rating in Figure \ref{fig:correlation_analysis}.

\subsection{Metric Evaluation in the Cited Evaluation Scenario}
\label{appendix:results_metric_eval_cited_only}

We further show the performance of different evaluation metrics in the \textbf{Cited} scenario in Table \ref{tab:results_metric_eval_star}. 
Consistent with the \textbf{Full} scenario, 
\textsc{CiteEval-Auto} metrics achieve superior correlation with human ratings compared existing methods.

\section{Potential Risks}
The efficiency of \textsc{CiteEval-Auto} carries the potential risk of over-reliance on automated assessments, potentially diminishing the critical role of human judgment in fully capturing the multifaceted aspects of citation quality. 
To counter this, it is crucial to emphasize that \textsc{CiteEval-Auto} is designed as a tool for efficient, scalable evaluation, not as a substitute for human expertise. CiteEval's principle-driven nature is intended to foster critical examination and iterative refinement, ultimately ensuring that human expertise remains central to the comprehensive assessment of citation quality, particularly in high-stakes applications.

\section{Prompt Templates}

\subsection{Prompt Template for RAG Response Generation with Citations}
\label{appendix:prompt_rag}

We show the prompt template for retrieval-augmented response generation with citations in Table \ref{tab:prompt_rag}.

\input{prompt_rag}

\subsection{Prompt Template for Context Attribution}
\label{appendix:prompt_statement_classification}

\input{prompt_context_attribution}

We show the prompt template for context attribution in Table \ref{tab:prompt_context_attribution}. 

\subsection{Prompt Template for Citation Editing and Rating}
\label{appendix:prompt_citation_editing}

We show the prompt template for joint citation editing and rating in Table \ref{tab:prompt_citation_editing}. 

\input{prompt_citation_editing}

\end{document}

%% file: fig_intro.tex
\begin{figure}[t]
\centering
\includegraphics[width=\linewidth]{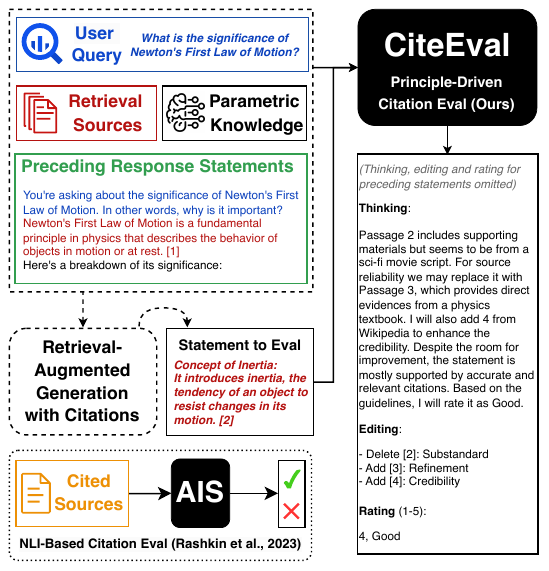} 

\caption{{CiteEval} considers all contexts used in the generation phase (with dashed lines) for fine-grained citation assessment. 
In contrast, AIS (\citealt{rashkin-etal-2023-ais}; bottom left) judges citation quality with NLI, solely based on \textcolor{myyellow}{cited sources} (e.g., Passage 2 in the shown example), a subset of \textcolor{myred}{retrieval sources}.
Statements in the response are in the same color as the contexts to which they should be attributed.}
\label{fig:intro} 
\end{figure}

%% file: tab_context_examples.tex
\begin{table}[t]
\small
\centering
\def \arraystretch{1.5}
\tabcolsep=0.1cm

\begin{tabularx}{\linewidth}{X}
\thickhline
\multicolumn{1}{c}{User Context}
\\

$\bullet$ Affirmation: \textit{Sometimes, when falling asleep, individuals may experience a heightened perception of sounds, which can feel extremely loud} when $Q$ asks about the reason behind sound sensitivity when falling asleep. 
\\

\hdashline
\multicolumn{1}{c}{Response Context} 
\\
$\bullet$ Mathematical reasoning: \textit{... Therefore, an Oreo without the filling has 21.6 calories}) \\
    
$\bullet$ Logical reasoning: \textit{... Therefore, it is not possible for some planets to orbit the Sun in the opposite direction} \\
    
$\bullet$ Planning: \textit{... However, I need to find who played Zordon in the original Power Rangers series} \\

$\bullet$ Abstention: \textit{I cannot find an answer from your documents.}
\\
\hdashline

\multicolumn{1}{c}{Parametric Context}
\\
$\bullet$ Factual: \textit{The molecular formula of phenol is C6H5OH} when the retrieval context only mentions its molar mass.  \\
    
$\bullet$ Formatting: \textit{The Washington Redskins went to the Super Bowl in the following years} for response coherence. 
\\

\thickhline
\end{tabularx}

\centering
\caption{Examples of statements attributable to contexts beyond retrieval.
}
\label{tab:context_examples}
\end{table}

%% file: tab_data_stats.tex
\begin{table}[t]
\small
\centering
\def \arraystretch{1.5}
\tabcolsep=0.08cm

\resizebox{\linewidth}{!}{
\begin{tabular}{llrr}
\thickhline
\textbf{Subset} & \textbf{Source} & \textbf{\#Instances} 
\\
\thickhline
ASQA & Wikipedia \typewriter{[W]}& 948 queries & \\

ELI5 & Sphere \typewriter{[S]} & 1,000 queries & \\

MS MARCO & Bing \typewriter{[B]} & 1,000 queries & \\

LFRQA & LoTTE \typewriter{[L]}  & 1,000 queries\\

Total & \typewriter{[W]}, \typewriter{[S]}, \typewriter{[B]}, \typewriter{[L]}  & 3,948 queries
\\

\hdashline
Full Development & \typewriter{[W]}, \typewriter{[S]}, \typewriter{[B]} & 948 queries
\\

Full Test & \typewriter{[W]}, \typewriter{[S]}, \typewriter{[B]}, \typewriter{[L]}  & 3,000 queries
\\

Metric Development & \typewriter{[W]}, \typewriter{[S]}, \typewriter{[B]} & 200 responses
\\
Metric Test & \typewriter{[W]}, \typewriter{[S]}, \typewriter{[B]} & 1,000 responses
\\
\thickhline
\end{tabular}
}

\vspace{1em}

\begin{tabularx}{\linewidth}{lX}
\thickhline
\textbf{Annot.} & \textbf{Ratio} \\
\thickhline
Contexts & 
retrieval (87.0\%), user (0.6\%), response (9.3\%), parametric (3.1\%)
\\
Ratings & 1-5: 10.3\%, 2.1\%, 8.6\%, 16.9\%, 62.0\% \\
Edits & 
\typewriter{delete}: {misleading} (6.9\%), {substandard} (1.3\%), {redundancy} (4.5\%);
% \newline
\typewriter{add}: evidence (11.3\%), {refinement} (1.4\%), {credibility} (6.7\%); 
% \newline
\typewriter{keep}: 67.8\%
\\
\thickhline
\end{tabularx}

\centering
\caption{{CiteBench} summary. 
We report the distributions of queries and responses (above) and human annotations for citation evaluation (below).
}

\label{tab:citebench_stats}
\end{table}

%% file: tab_human_correlation.tex
% left block: sentence
% right block: response
\begin{table*}[ht]
\resizebox{\textwidth}{!}{
\small
\centering
\def \arraystretch{1.5}
\tabcolsep=0.14cm
\begin{tabular}{llrrrrrr}
\thickhline
\multicolumn{2}{c}{Evaluator} & \multicolumn{3}{c}{{CiteBench-Statement}} & \multicolumn{3}{c}{{CiteBench-Response}}\\
\textbf{Metric} & \textbf{Model} & \textbf{Pearson} & \textbf{Spearman} & \textbf{Kendall-Tau} & \textbf{Pearson} & \textbf{Spearman} & \textbf{Kendall-Tau}\\
\thickhline
\multicolumn{8}{c}{\textit{AutoAIS-based Metrics}}
\\

\textsc{AutoAis-Precision} 
% \cite{gao-etal-2023-alce} 
& T5-XXL 
& ------ & ------ & ------
& 0.170 & 0.058 & 0.057
\\

\textsc{AutoAis-Recall} 
% \cite{gao-etal-2023-alce} 
& T5-XXL 
& 0.409 & 0.264 & 0.237
& 0.223 & 0.075 & 0.073
\\

\textsc{AutoAis-F1} 
% \cite{gao-etal-2023-alce} 
& T5-XXL 
& ------ & ------ & ------
& 0.219 & 0.105 & 0.097
\\

\textsc{AutoAis-Precision}$^\dagger$ & T5-XXL 
& 0.416 & 0.315 & 0.278
& 0.256 & 0.113 & 0.106
\\

\textsc{AutoAis-F1}$^\dagger$ & T5-XXL 
& 0.419 & 0.315 & 0.278
& 0.249 & 0.115 & 0.108
\\
\hdashline

\multicolumn{8}{c}{\textit{AttriScore-based Metrics}}
\\

\textsc{AttriScore-Strict}$^\dagger$
% \cite{yue-etal-2023-attrscore} 
& GPT-4-turbo
& 0.459 & 0.281 & 0.254
& 0.196 & 0.079 & 0.097
\\

\textsc{AttriScore-Relaxed}$^\dagger$
% \cite{yue-etal-2023-attrscore}
& GPT-4-turbo
& 0.447 & 0.274 & 0.249
& 0.098 & 0.066 & 0.092
\\

\textsc{AttriScore-Strict}$^\dagger$
% \cite{yue-etal-2023-attrscore} 
& GPT-4o 
& 0.449 & 0.297 & 0.269
& 0.221 & 0.094 & 0.108
\\

\textsc{AttriScore-Relaxed}$^\dagger$
% \cite{yue-etal-2023-attrscore} 
& GPT-4o 
& 0.450 & 0.291 & 0.263
& 0.128 & 0.080 & 0.104
\\

\hdashline

\multicolumn{8}{c}{\textit{LQAC-based Metrics}}
\\

\textsc{LQAC-Precision} 
% \cite{zhang2024longcite} 
& GPT-4o 
& ------ & ------ & ------
& -0.043 & -0.092 & -0.057
\\

\textsc{LQAC-Recall} 
% \cite{zhang2024longcite}
& GPT-4o 
& 0.607	& 0.423	& 0.375
& 0.526	& 0.447	& 0.379 
\\

\textsc{LQAC-F1} 
% \cite{zhang2024longcite}
& GPT-4o 
& ------ & ------ & ------
& 0.118 & 0.071 & 0.090
\\

\textsc{LQAC-Precision}$^\dagger$ & GPT-4o 
& 0.468 & 0.269 & 0.247
& 0.147 & 0.052 & 0.082
\\

\textsc{LQAC-F1}$^\dagger$ 
& GPT-4o
& 0.468 & 0.284 & 0.255
& 0.182 & 0.074 & 0.096
\\

\hdashline

\multicolumn{8}{c}{\textit{CiteEval-Auto Metrics (Ours)}}
\\

\textsc{CiteEval-Auto} (\textsc{IterCoE}) & GPT-4o 
& 0.710	& 0.549	& \textbf{0.491}
& 0.647	& 0.580	& \textbf{0.499}
\\

\textsc{CiteEval-Auto} (\textsc{EditDist}) & GPT-4o+MLR 
&0.711	&0.558	&0.486
&0.633	&0.585	&0.487
\\

\textsc{CiteEval-Auto} & GPT-4o+MLR 
& \textbf{0.731}	& \textbf{0.559}	&0.486
& \textbf{0.668}	& \textbf{0.589}	&0.492
\\

\thickhline
\end{tabular}
}
\centering
\caption{Human correlation of different citation evaluation metrics on CiteBench (metric test set). 
$\dagger$~denotes our adapted version
described in Section \ref{subsec:metric_eval_setup} for  statement-level evaluation.
\textsc{CiteEval-Auto} (last row) is an ensemble which linearly interpolates scores from the two proposed rating methods. 
}
\label{tab:results_metric_eval}
\end{table*}

%% file: fig_ablation.tex
\begin{figure}[t]
  \centering % Centers the figure
  \begin{subfigure}{0.23\textwidth} % First subfigure, 45% of the text width
    \centering
    \includegraphics[width=\textwidth]{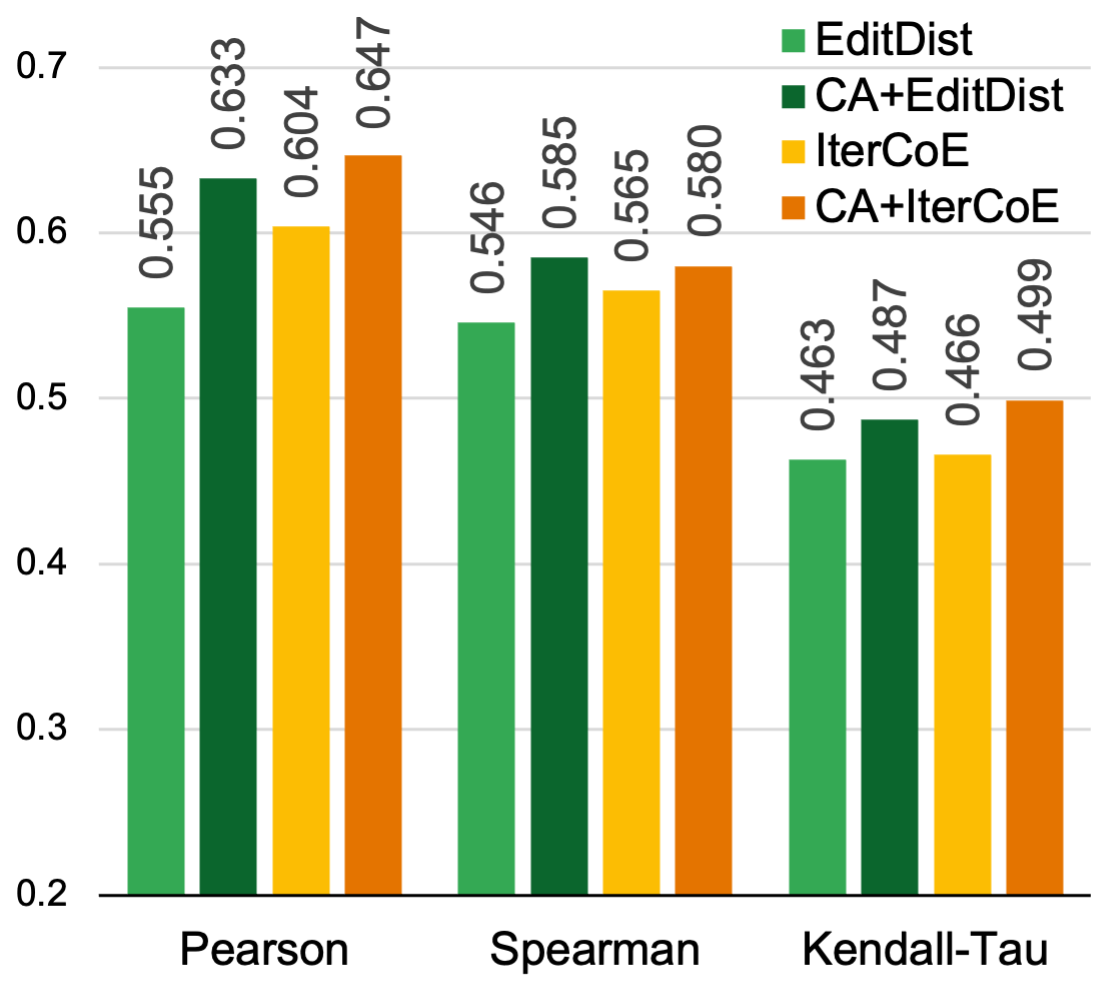} 
    % \caption{Correlation between response length and citation density.} 
    \label{fig:image1}
  \end{subfigure}
  \hfill % Adds horizontal space between the subfigures
  \begin{subfigure}{0.23\textwidth} % Second subfigure, 45% of the text width
    \centering
    \includegraphics[width=\textwidth]{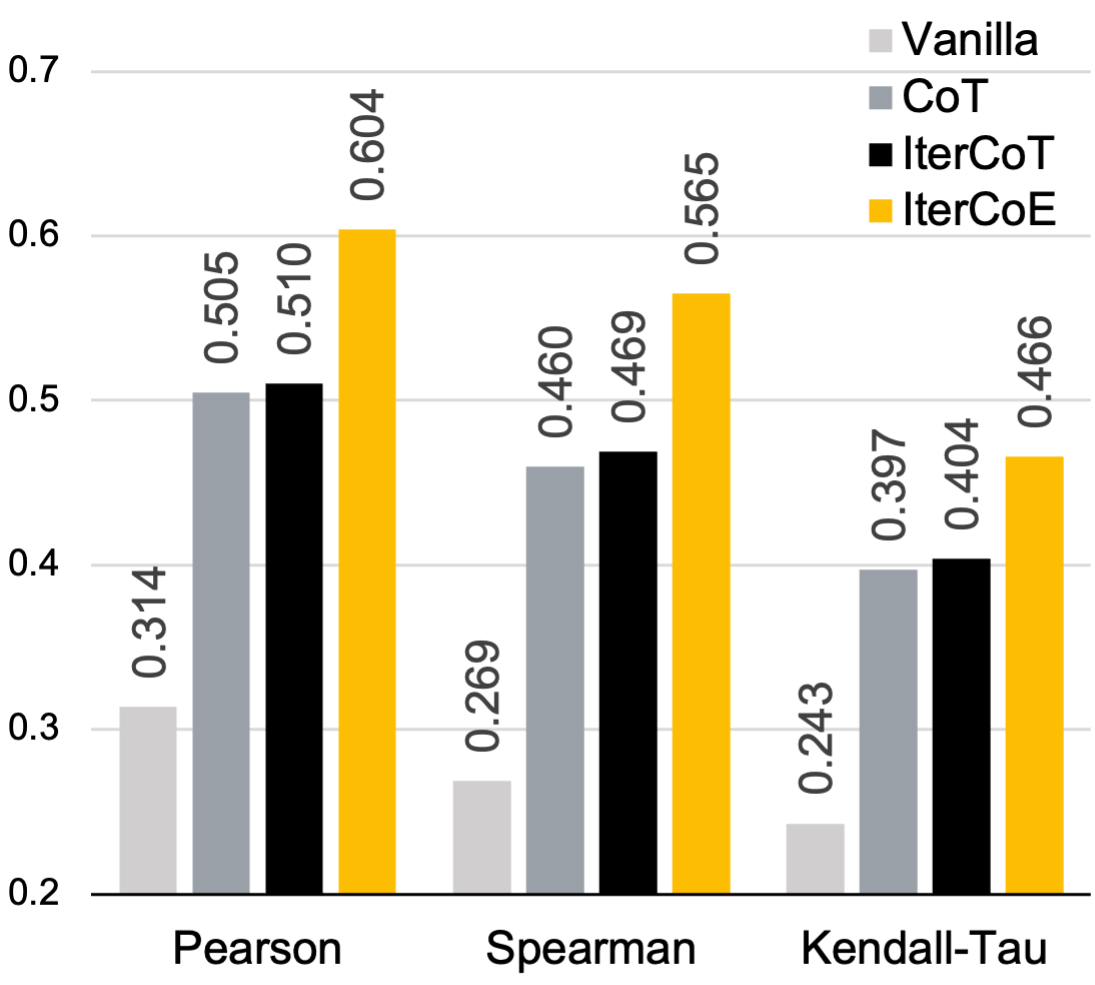}
    % \caption{Correlation between citation density and citation rating.} 
    \label{fig:image2}
  \end{subfigure}
  \caption{Effects of context attribution (left) and citation editing (right) in citation evaluation.}
  \label{fig:citeeval_ablation}
\end{figure}

%% file: fig_estimated_edit_dist.tex
\begin{figure}[t]
\centering
\includegraphics[width=0.46\textwidth]{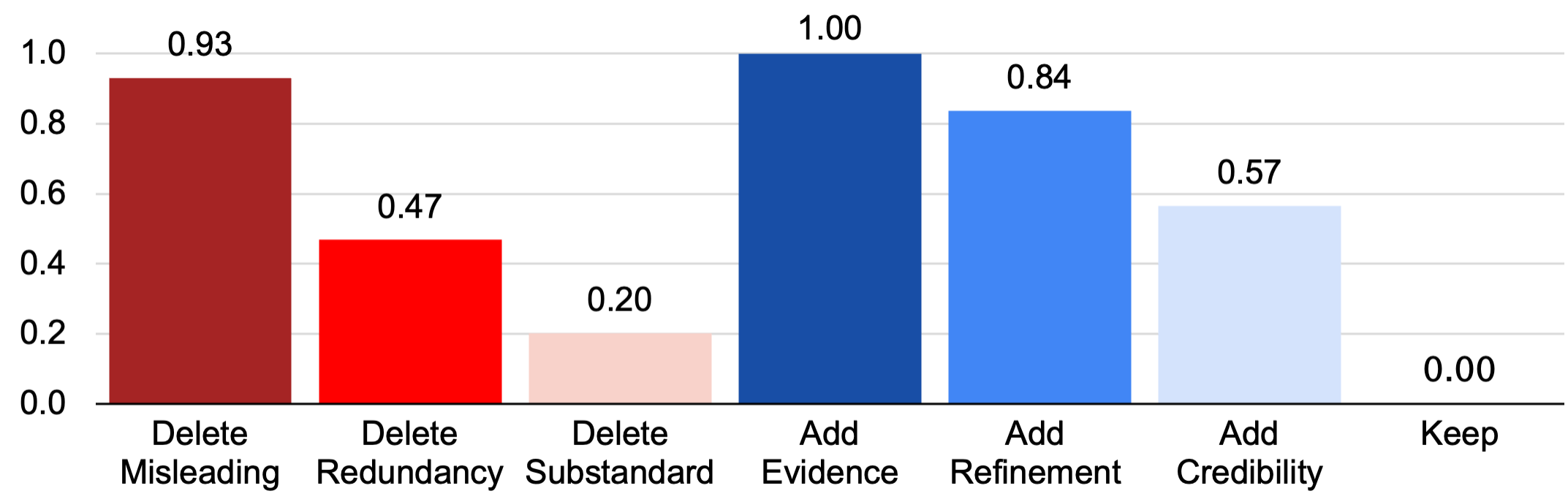} 
\caption{Edit distance for actions in \textsc{EditDist}. Estimated on the metric development set.}
\label{fig:estimated_edit_dist} 
\end{figure}

%% file: tab_rag_results.tex
\begin{table}[t]
\resizebox{\linewidth}{!}{
\small
\centering
\def \arraystretch{1.5}
\tabcolsep=0.3cm
\begin{tabular}{lcccc}
\thickhline
\multirow{2}{*}{Models} & \multicolumn{2}{c}{\textsc{CiteEval-Auto}} & \multicolumn{2}{c}{Statistics} \\
 & \textbf{Full} & \textbf{Cited} &
$\bm{|R|}$ & $\bm{M}$ \\
\thickhline
\multicolumn{5}{c}{\textit{Proprietary Models}}
\\
GPT-4o	
&0.898	&\textbf{0.949}	&1.975	&0.197
\\

GPT-4o-mini	
&0.848	&0.925	&1.759	&0.217
\\

GPT-4-turbo	
&0.863	&0.940	&1.835	&0.250
\\

GPT-3.5-turbo
&0.724	&0.839	&1.352	&0.223
\\

\hline

\multicolumn{5}{c}{\textit{Open-source Instruction-tuned Models}}
\\

Llama-3-70b	
&\textbf{0.909}	&0.926 &1.853	&0.159

\\

Llama-3-8b
&0.800	&0.871	&2.398	&0.250
\\

Mixtral-8$\times$22b
&0.746	&0.871	&2.322	&0.386
\\

Mixtral-8$\times$7b 
&0.755	&0.827	&2.554	&0.363
\\

Qwen2.5-72b 
& 0.895 & 0.913 & 1.461 & 0.161
\\

Qwen2.5-7b 
& 0.663 & 0.722 & 8.467 & 0.950
\\

\hline
\multicolumn{5}{c}{\textit{Open-source Fine-tuned Models}}
\\

LongCite-9B 
& 0.564 & 0.843 & 8.867 & 0.435
\\

LongCite-8B 
& 0.559 & 0.846 & 8.694 & 0.452

\\
\thickhline
\end{tabular}
}
\centering
\caption{Citation quality in \textbf{Full} and \textbf{Cited} scenarios of different LLMs responses (full test set). 
We also report the response length $|R|$ and missing citation ratio $M$.
}
\label{tab:rag_results}
\end{table}

%% file: fig_effects_of_retrieval.tex
\begin{figure}[t]
\centering
\includegraphics[width=0.46\textwidth]{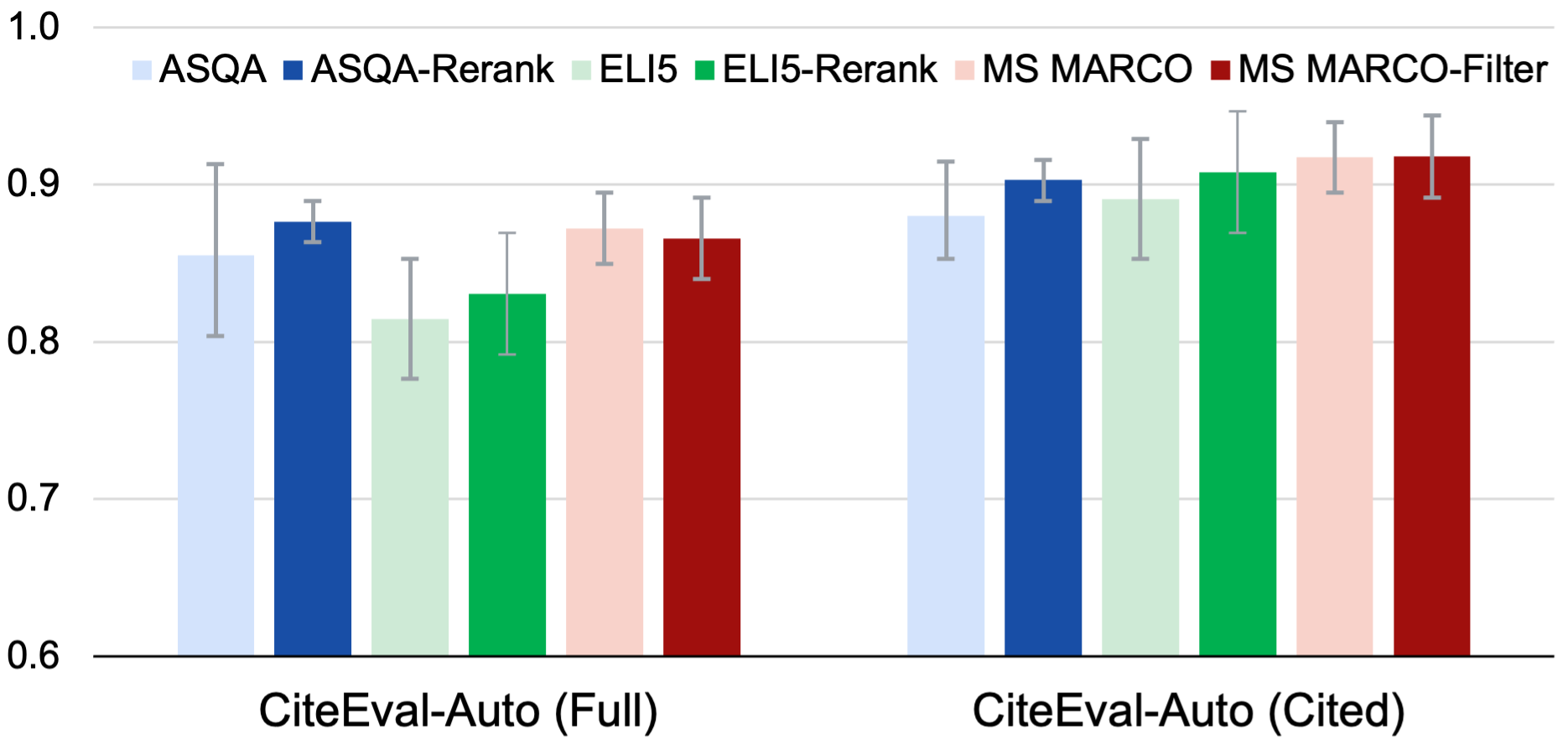} 
\caption{Performance of different retrieval settings (full development set) in both \textbf{Full} and \textbf{Cited} scenarios. Error bars denote the standard deviation over averaged ratings for different models.}
\label{fig:effects_of_retrieval} 
\end{figure}

%% file: fig_iterative_improvement.tex
\begin{figure}[t]
\centering
\includegraphics[width=0.46\textwidth]{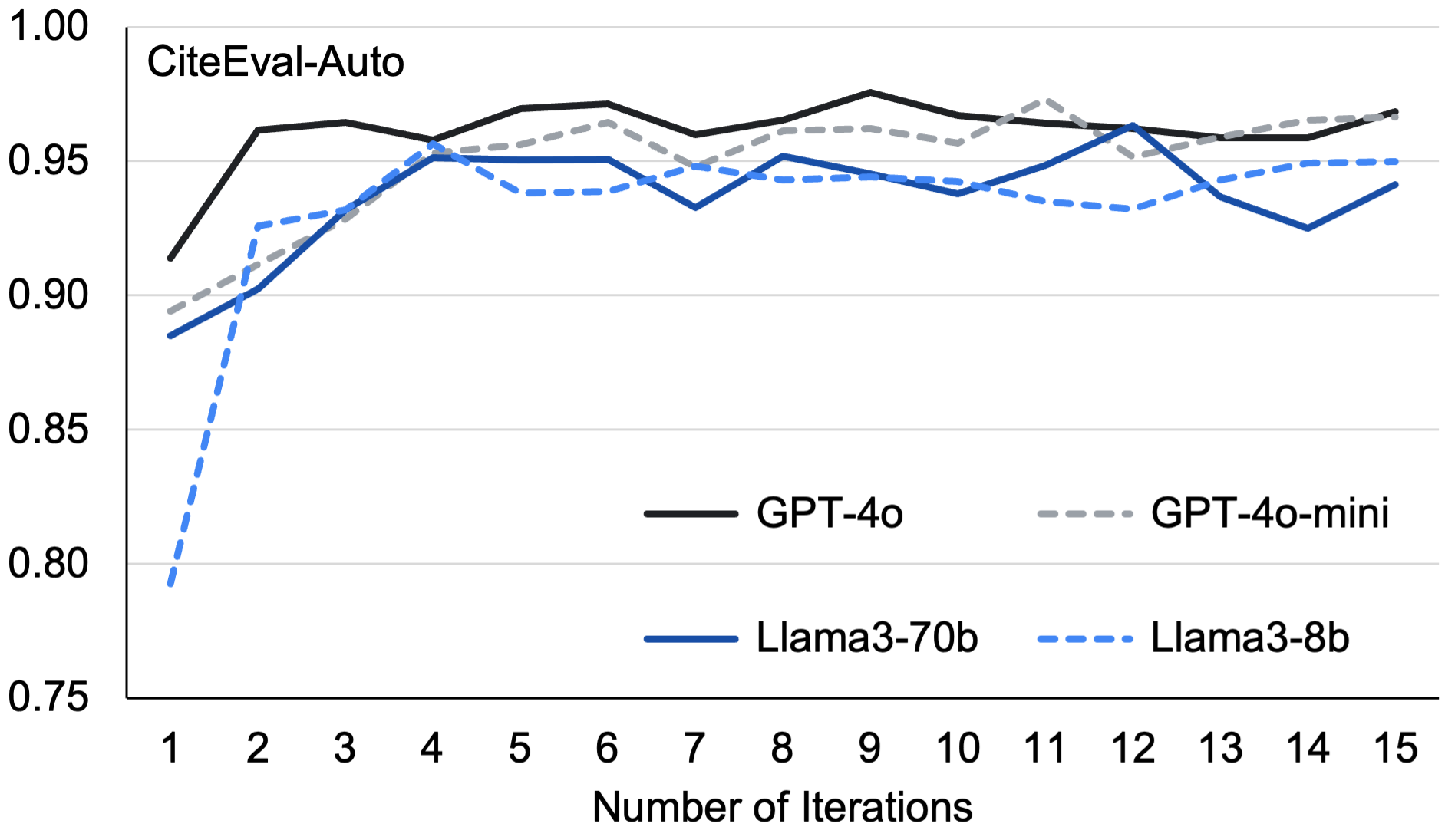} 
\caption{Performance improvement with iterative citation editing  (response-level; metric development set).}
\label{fig:iterative_improvement} 
\end{figure}

%% file: tab_edit_description.tex
\begin{table*}[ht]
\small
\centering
\def \arraystretch{1.5}
\tabcolsep=0.1cm
\begin{tabularx}{\textwidth}{lX}
\thickhline
\textbf{Edit} &\textbf{Description}\\
\thickhline
\typewriter{delete}-mislead
& Irrelevant citation. Removing this citation can avoid misleading users. \\
\typewriter{delete}-substandard & Relevant citation, however another source is more helpful and should be cited instead.\\
\typewriter{delete}-redundancy
&  Relevant citation, however other citations (for the same statement or a larger cited context) contain sufficient supporting evidence.  Removing this citation can improve conciseness.\\

\hdashline

\typewriter{add}-evidence  & Existing citations lack certain required evidence, leaving the statement partially or fully unsupported. Adding this citation can fill the gap with the required evidence.\\
\typewriter{add}-refinement & An existing citation is relevant but with suboptimal source quality. This new source is more helpful and should be cited instead (an existing citation should be deleted). \\
\typewriter{add}-credibility & Existing citations cover all essential evidence from optimal sources. Adding the citation can further enhance  response credibility. \\
\thickhline
\end{tabularx}
\centering
\caption{Citation edit actions in CiteEval and their applicable scenarios.
}
\label{tab:edits}
\end{table*}

%% file: fig_response_token_dist.tex
\begin{figure}[t]
\centering
\includegraphics[width=0.46\textwidth]{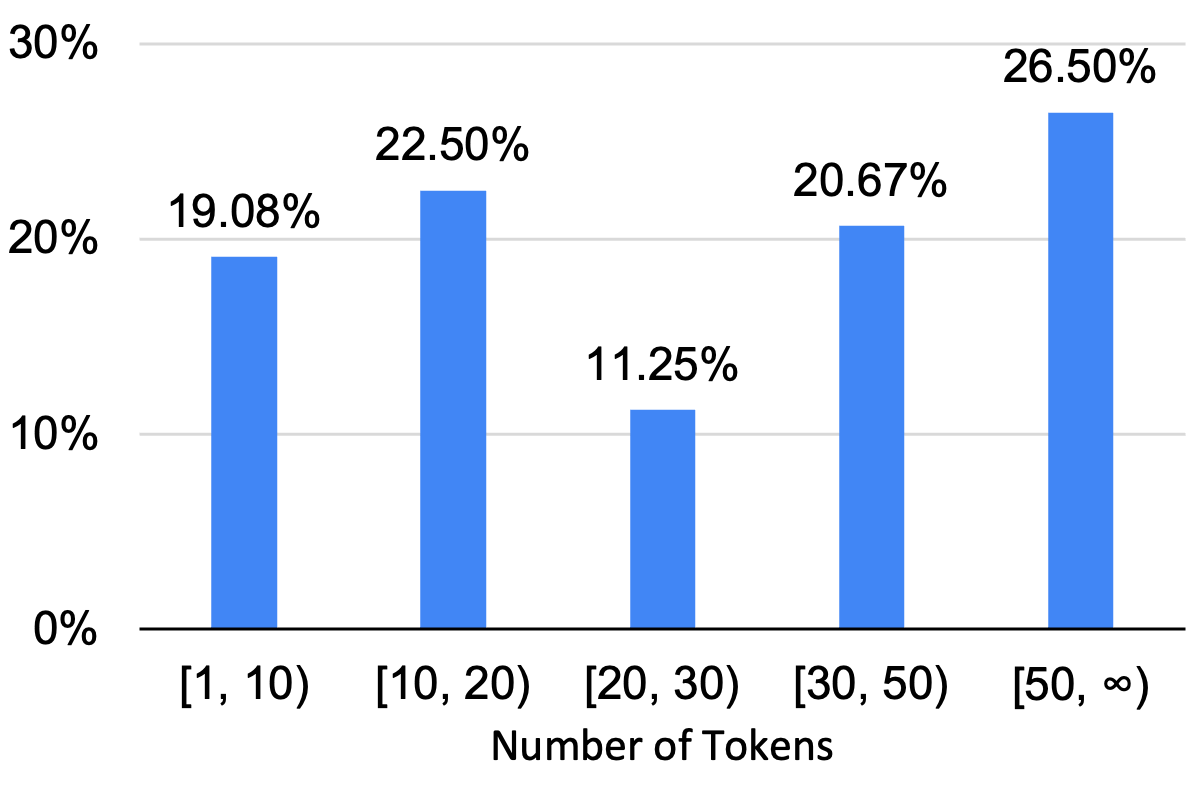} 
\caption{Length distribution of model responses in CiteBench.}
\label{fig:token_dist} 
\end{figure}

%% file: tab_results_citeeval_llms.tex
% left block: sentence
% right block: response
\begin{table}[t]
\small
\centering
\def \arraystretch{1.5}
\tabcolsep=0.15cm
\begin{tabular}{lccc}
\thickhline
\textbf{Model} & \textbf{Pearson} & \textbf{Spearman} & \textbf{Kendall-Tau} \\
\thickhline

\multicolumn{4}{c}{\textit{CiteBench-Statement}}\\

% \hdashline
GPT-4o 
& \textbf{0.731}	& \textbf{0.559}	&\textbf{0.486} \\

GPT-4-turbo & 0.721	& 0.513	& 0.444 \\

\hdashline

\multicolumn{4}{c}{\textit{CiteBench-Response}}\\
% \hdashline
GPT-4o  & \textbf{0.668}	& \textbf{0.589}	&\textbf{0.492} \\

GPT-4-turbo & 0.647	& 0.546	& 0.454 \\

\thickhline
\end{tabular}
\centering
\caption{Human correlation of different LLMs for \textsc{CiteEval-Auto} (metric test set).
}
\label{tab:citeeval_llms}
\end{table}

\begin{table}[t]
\small
\centering
\def \arraystretch{1.5}
\tabcolsep=0.3cm
\begin{tabular}{lccc}
\thickhline
\textbf{Contexts} & \textbf{Precision} & \textbf{Recall} & \textbf{F1} \\
\thickhline
Applicable & 0.992 & 0.988 & 0.990\\
Not Applicable & 0.910 & 0.937 & 0.923\\
Average & 0.951 & 0.962 & 0.957 \\
\thickhline
\end{tabular}
\caption{Performance of model-based context attribution in \textsc{CiteEval-Auto} (metric development set).}
\label{tab:context_attribution_results}
\end{table}

\begin{figure}[t]
\centering
\includegraphics[width=0.5\textwidth]{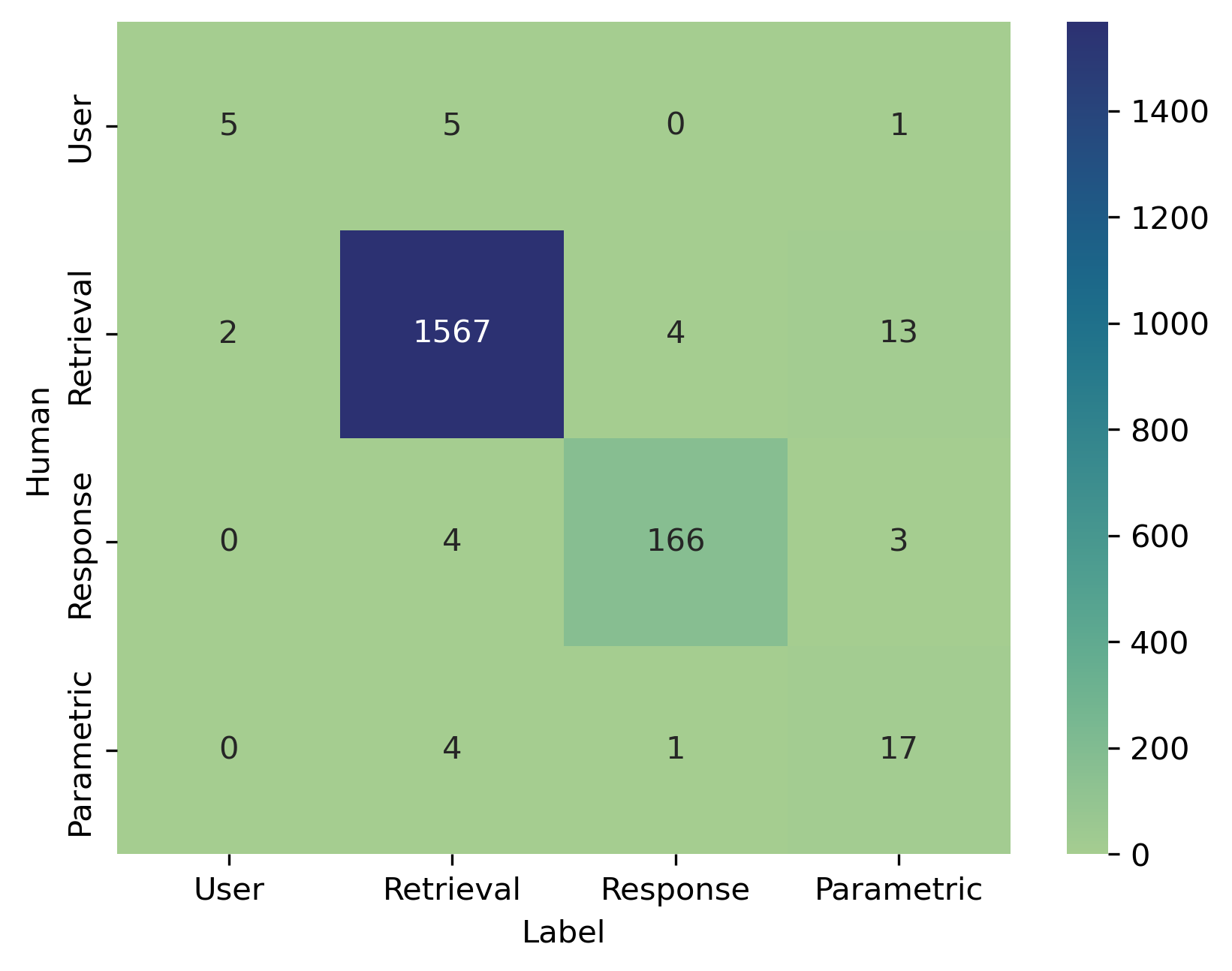} 
% \vspace{-1.5em}
\caption{Confusion matrix for \textsc{CiteEval-Auto} context attribution (metric development set).}
\label{fig:conf_mat} 
\end{figure}

%% file: tab_human_correlation_star.tex
% left block: sentence
% right block: response
\begin{table*}[ht]
\small
\centering
\def \arraystretch{1.5}
\tabcolsep=0.15cm
\begin{tabular}{llrrrrrr}
\thickhline
\multicolumn{2}{c}{Evaluator} & \multicolumn{3}{c}{\textsc{CiteBench-Statement}} & \multicolumn{3}{c}{\textsc{CiteBench-Response}}\\
\textbf{Metric} & \textbf{Model} & \textbf{Pearson} & \textbf{Spearman} & \textbf{Kendall-Tau} & \textbf{Pearson} & \textbf{Spearman} & \textbf{Kendall-Tau}\\
\thickhline
\multicolumn{8}{c}{\textit{AutoAIS-based Metrics}}
\\

\textsc{AutoAis-Precision} 
% \cite{gao-etal-2023-alce} 
& T5-XXL 
& ------ & ------ & ------
& 0.187 & 0.065 & 0.062
\\

\textsc{AutoAis-Recall} 
& T5-XXL 
& 0.227 & 0.136 & 0.122
& 0.119 & -0.022 & -0.014
\\

\textsc{AutoAis-F1} 
& T5-XXL 
& ------ & ------ & ------
& 0.155 & 0.038 & 0.039
\\

\textsc{AutoAis-Precision}$^\dagger$ & T5-XXL 
& 0.268 & 0.209 & 0.184
& 0.181 & 0.048 & 0.047
\\

\textsc{AutoAis-F1}$^\dagger$ & T5-XXL 
& 0.253 & 0.202 & 0.178
& 0.153 & 0.032 & 0.034
\\
\hdashline

\multicolumn{8}{c}{\textit{AttriScore-based Metrics}}
\\

\textsc{AttriScore-Strict}$^\ast$
& GPT-4-turbo
& 0.459 & 0.281 & 0.254
& 0.196 & 0.079 & 0.097
\\

\textsc{AttriScore-Relaxed}$^\ast$
& GPT-4-turbo
& 0.447 & 0.274 & 0.249
& 0.098 & 0.066 & 0.092
\\

\textsc{AttriScore-Strict}$^\ast$
% \cite{yue-etal-2023-attrscore} 
& GPT-4o 
& 0.449 & 0.297 & 0.269
& 0.221 & 0.094 & 0.108
\\

\textsc{AttriScore-Relaxed}$^\ast$
& GPT-4o 
& 0.450 & 0.291 & 0.263
& 0.128 & 0.080 & 0.104
\\

\hdashline

\multicolumn{8}{c}{\textit{LQAC-based Metrics}}
\\

\textsc{LQAC-Precision} 
& GPT-4o 
& ------ & ------ & ------
& -0.011 & -0.079 & -0.046
\\

\textsc{LQAC-Recall} 
& GPT-4o 
& 0.329 & 0.275 & 0.241
& 0.338 & 0.290 & 0.245
\\

\textsc{LQAC-F1} 
& GPT-4o 
& ------ & ------ & ------
& 0.022 & -0.037 & -0.011
\\

\textsc{LQAC-Precision}$^\dagger$ & GPT-4o 
& 0.137 & 0.093 & 0.086
& 0.020 & -0.080 & -0.049
\\

\textsc{LQAC-F1}$^\dagger$ 
& GPT-4o
& 0.174 & 0.130 & 0.117
& 0.033 & -0.055 & -0.027
\\

\hdashline

\multicolumn{8}{c}{\textit{CiteEval-Auto Metrics (Ours)}}
\\

\textsc{CiteEval-Auto} (\textsc{IterCoE}) & GPT-4o 
& 0.464 & 0.432 & \textbf{0.383}
& 0.501 & 0.472 & \textbf{0.404}
\\

\textsc{CiteEval-Auto} (\textsc{EditDist}) & GPT-4o+MLR 
& 0.397 & 0.435 & 0.374
& 0.431 & 0.472 & 0.389
\\

\textsc{CiteEval} & GPT-4o+MLR 
& \textbf{0.469} & \textbf{0.441} & 0.378
& \textbf{0.502} & \textbf{0.482} & 0.397
\\

\thickhline
\end{tabular}
\centering
\caption{Human correlation of different evaluation metrics in the \textbf{Cited} scenario (metric test set).}
\label{tab:results_metric_eval_star}
\end{table*}

%% file: fig_correlation.tex
\begin{figure}[t]
  \centering

  \begin{subfigure}{0.48\columnwidth} % First subfigure, 45% of the text width
    \centering
    \includegraphics[width=\columnwidth]{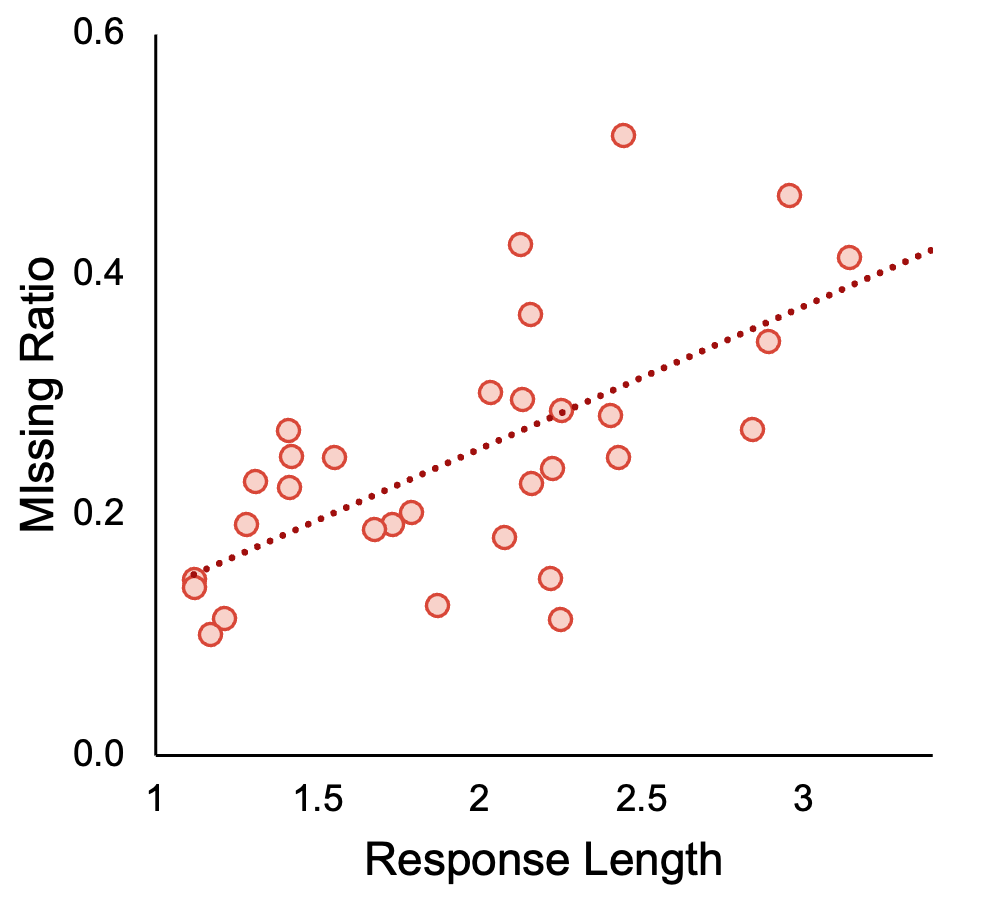} 
    % \caption{Correlation between response length and citation density.} 
    \label{fig:image1}
  \end{subfigure}
  \hfill % Adds horizontal space between the subfigures
  \begin{subfigure}{0.48\columnwidth} % Second subfigure, 45% of the text width
    \centering
    \includegraphics[width=\columnwidth]{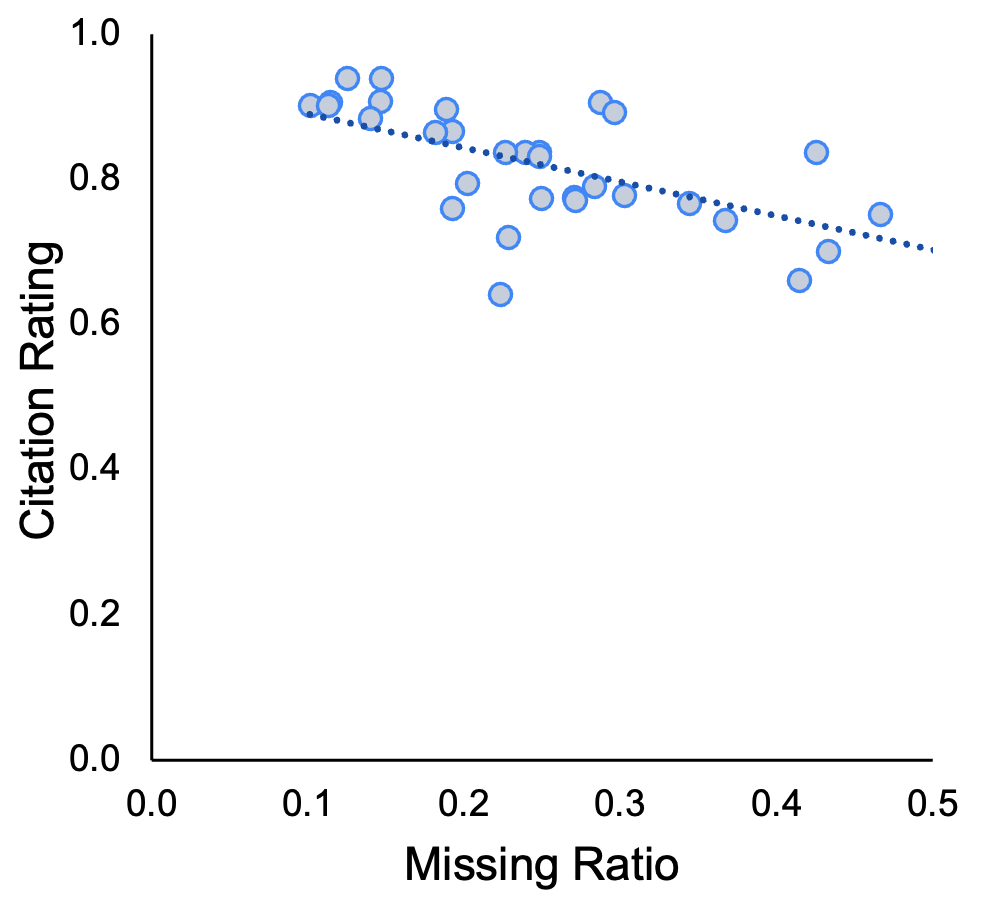}
    % \caption{Correlation between citation density and citation rating.} 
    \label{fig:image2}
  \end{subfigure}
  \vspace{-1em}
  \caption{Correlation analysis between response length and missing citation ratio (left; Pearson correlation $0.679$, $p<.001$), and missing citation ratio and citation rating (right; Pearson correlation $-0.633$, $p<.001$). 
  Each data point represents the averaged results for one model-dataset pair from the full CiteBench test set.} % Overall caption for the entire figure
  \label{fig:correlation_analysis} % Label for referencing the entire figure later
\end{figure}

%% file: prompt_rag.tex
\begin{table*}[ht]
\tt
\noindent
\begin{tcolorbox}[colback=black!5!white,colframe=black!75!black,title=RAG Response Generation with Citations]
\small
\begin{minipage}{\textwidth} 
Provide an answer to the question using information from the given passages. Passages are provided inside the <passage> </passage> XML tags. Question is provided inside the <question> </question> XML tags.\newline
\newline
Add passage id in brackets at the end of each answer sentence to cite passages in <passage> for any factual claim. Don't use "[passage [1]]" when citing. Instead use solely passage id in brackets such as [1]. When citing several passages, use [1][2][3]. For each sentence in your answer that contains factual claims, cite at least one passage and at most three passages.\newline
\newline
Below are the passages. Each passage has an id for citation:\newline
\newline
[[Retrieved Passages Go Here]]\newline
\newline
Below is the question:\newline
\newline
<question>\newline
\newline
[[User Query Goes Here]]\newline
\newline
</question>\newline
\newline
Now answer the question using only information from the passages. Think step by step first and put your thinking process into <thinking> </thinking> tags. The thinking process should not exceed 50 words. Provide the final answer after the thinking process. Remember to do citation for the final answer using bracketed numbers at sentence end. In your final answer, do not use expressions similar to "Passage 1", "Passage [1]", "according to Passage [1]" to show your thought process or justify your citations in your answer. Remember that you need to say "No answer is found" if the question cannot be answered by information in the passages.
\end{minipage}
\end{tcolorbox}
\caption{Prompt template for RAG response generation with citations.}
\label{tab:prompt_rag}
\end{table*}

%% file: prompt_context_attribution.tex
\begin{table*}[ht]
\centering
\tt
\noindent
\begin{tcolorbox}[colback=black!5!white,colframe=black!75!black,title=Context Attribution]
\begin{minipage}{\textwidth} 
\small
You are an expert specializing in analyzing sentences within a given model response and classifying them based on their attribution. \newline

You task is to carefully examine each sentence, and attribute it to one of the following categories:\newline

<categories>\newline
1. Query: Sentences that iterate or rephrase the user query without making new claims or involving new facts.\newline

2. Retrieval: Sentences fully or partially supported by the retrieval context.\newline

3. Response: Sentences solely derived from preceding sentences within the response itself, not relying on the query context, the retrieval context, or the succeeding sentences in the response. Examples include sentences that perform mathematical and logical reasoning over preceding response sentences. \newline

4. Model: Sentences solely based on the inherent knowledge of the language model that generated the response. Knowledge is only inherent when it can NOT be found in, or reasonably inferred from, the query context, the retrieval context, or the response context. Examples include unsupported facts, and transitional expressions/summarization without any substantial claims.\newline
<categories>\newline

Follow the guidelines below for ambiguous cases: \newline

<ambiguous\_cases>\newline
- For sentences involving both the retrieval context and other types of contexts, choose 2 (Retrieval).\newline
- For single-sentence responses indicating that no answer could be found, choose 3 (Response).\newline
- For sentences supported by its succeeding sentences but not its preceding sentences, choose from 1 (Query), 2 (Retrieval) and 4 (Model).\newline
</ambiguous\_cases>

Below is the query:\newline

<query>\newline
[[User Query]]\newline
</query>\newline

Below is the retrieval context, consisting of documents retrieved for the query:\newline

<retrieval>\newline
[[Retrieved Passages]]\newline
</retrieval>\newline

Below is the response, consisting of the sentences to evaluate:\newline

<response>\newline
[[Response Sentences Go Here]]\newline
</response>\newline

From now on you must follow this format:\newline

<thinking> Think step by step first before classifying sentence 1 </thinking>\newline
<category sentence\_id="1"> Choose the attribution of sentence 1 from 1, 2, 3, 4 </category>\newline
<thinking> Think step by step first before classifying sentence 2 </thinking>\newline
<category sentence\_id="2"> Choose the attribution of sentence 2 from 1, 2, 3, 4 </category>\newline
...\newline
<thinking> Think step by step first before classifying sentence N </thinking>\newline
<category sentence\_id="N"> Choose the attribution of sentence N from 1, 2, 3, 4 </category>\newline

Begin!
\end{minipage}
\end{tcolorbox}
\caption{Prompt template for context attribution in \textsc{CiteEval-Auto}.}
\label{tab:prompt_context_attribution}
\end{table*}

%% file: prompt_citation_editing.tex
\begin{table*}[ht]
\centering
\tt

\noindent
\vspace{-5em}
\begin{tcolorbox}[colback=black!5!white,colframe=black!75!black,title=Citation Editing and Rating]
\begin{minipage}{\textwidth} 
\small
You are an expert specializing in analyzing, editing, and rating citations for sentences within a given model response.\newline

Your task is to carefully examine the citations for each sentence, provide critical editing to the citations, and rate the citation quality.\newline

You are allowed to use a sequence of DELETE or ADD edits for critical editing. Each edit operates on one citation.\newline

<edits>\newline
DELETE: You can delete a citation due to the following reasons:\newline
    DELETE REASON 1. Misleading: the citation is irrelevant, and removing this citation avoids misleading users.\newline
    DELETE REASON 2. Substandard: the citation is relevant, however another source is more helpful and should be cited instead.\newline
    DELETE REASON 3. Redundant: the citation is relevant, however other citations contain sufficient supporting evidence. Removing this citation improves conciseness.\newline

ADD: You should only add a citation due to the following reasons:\newline
    ADD REASON 1. Evidence: existing citations lack certain required evidence, leaving the statement partially or fully unsupported. Adding this citation fills the gap with the required evidence.\newline
    ADD REASON 2. Refinement: an existing citation is relevant but substandard. This new source is more helpful and should be cited instead (an existing citation should be deleted).\newline
    ADD REASON 3. Credibility: existing citations cover all essential evidence from optimal sources. Adding this citation further enhances response credibility.\newline
</edits>\newline

Each edit should be passed in as <{{edit\_name}} citation="\{\{citation\}\}">\{\{reason\}\}<\{\{edit\_name\}\}>, where {{edit\_name}} is the name of the specific edit (DELETE or ADD), \{\{citation\}\} is a citation id to be deleted or added, and \{\{reason\}\} is one of the reasons from <edits></edits>.\newline

You should replace \{\{edit\_name\}\}, \{\{citation\}\} and \{\{reason\}\} with the appropriate value.\newline

Below are the editing guidelines. Follow the guidelines when deciding whether and how to perform an edit.\newline

<editing\_guidelines>\newline
- Use N/A if no editing is needed.\newline
- Add 0 as the citation id for facts that can NOT be found in, or reasonably inferred from, the query context, the retrieval context, or the response context. This attributes the unsupported facts to inherent knowledge of the language model that generated the response.\newline
- You should aim to achieve citations of the highest standard with minimal editing. After editing, all major claims in the statement should be cited.\newline
- After editing, the citations should cite sources that are mostly helpful, when there are multiple related sources. The final citations for each sentence typically contain at most 3 citations, but there can be exceptions.\newline
</editing\_guidelines>\newline

After providing edits, rate the original citations for each sentence, following the guidelines below: \newline

<rating\_guidelines>\newline
- 5 (Excellent): The sentence is fully supported by all relevant and accurate citations. There are no unnecessary, misleading, or missing citations. The citations (if present) enhance the credibility and informativeness of the sentence.\newline
- 4 (Good): The sentence is mostly supported by accurate and relevant citations. One potentially relevant citation may be missing, or a slightly unnecessary citation may be present, but these do not significantly detract from the overall quality of the sentence.\newline
- 3 (Fair): The sentence has some issues with citations. There might be one or few noticeable missing citation that somewhat weaken the sentence's support, or there might be several unnecessary or inaccurate citations that detract from the sentence's clarity or conciseness. Overall, the sentence's accuracy and credibility are somewhat compromised.\newline
- 2 (Poor): The sentence has significant problems with citations. There might be multiple missing citations that leave that leave central claims unsupported, or there might be multiple unnecessary or inaccurate citations that significantly undermine the sentence's accuracy and credibility.\newline
- 1 (Unacceptable): The sentence is completely unsupported by citations or is supported entirely by inaccurate, irrelevant, or misleading citations. The sentence is rendered misleading and unreliable.\newline
</rating\_guidelines>
\end{minipage}
% }
\end{tcolorbox}
\caption{Prompt template for citation editing and rating in \textsc{CiteEval-Auto}.}
\label{tab:prompt_citation_editing}
% \end{longtable}
\end{table*}

\begin{table*}[ht]
\centering
\tt
\noindent
% \fbox{
\begin{tcolorbox}[colback=black!5!white,colframe=black!75!black,title=Citation Editing and Rating]
\begin{minipage}{\textwidth} 
\small
Below is a hypothetical example. \newline

<example> \newline
Given 10 passages related to the question "Can you explain the concept of time dilation in the context of special relativity?", and a response which has the following sentence and citations:
<citation sentence\_id="1", sentence="Time dilation occurs because the speed of light in a vacuum is constant for all observers, regardless of their relative motion."> 1, 6 </citation>\newline

The following example shows how you should improve the citations for this sentence:\newline

<thinking> This claim is directly supported by passage 1. However, passage 6 does not provide any direct evidence to the question, so I should delete it to avoid misleading users. Additionally, passage 7 clearly states that time dilation occurs due to the constant speed of light in a vacuum. It will constitute to a good citation, so I will add 7 for credibility. Based on these edits, I will rate the given citations 2 (Poor). </thinking> \newline
<editing sentence\_id="1">\newline
<DELETE citation="6"> DELETE REASON 1 </DELETE>\newline
<ADD citation="7"> ADD REASON 3 </ADD>\newline
</editing>\newline
<rating sentence\_id="1"> 2 </rating>\newline
</example>\newline

Below is the query:\newline

<query>\newline
[[User Question Goes Here]]\newline
</query>\newline

Below are the retrieved sources. Each source passage <passage> </passage> has an id for citation. \newline

<retrieval>\newline
[[Retrieved Passages Go Here]]\newline
</retrieval>\newline

Below is the response:\newline

<response>\newline
[[Response Goes Here]]\newline
</response>\newline

Below are the citations to evaluate. Each <citation> has a response sentence and its sentence id that it cites for.\newline

<citations>\newline
[[Citations Go Here]]\newline
</citations>\newline

From now on you must follow this format:\newline

<thinking> Think step by step first before editing citations for sentence 1. </thinking>\newline
<editing sentence\_id="1"> edits for citations in sentence 1, or N/A if no editing is needed </editing>\newline
<rating sentence\_id="1"> rating for citations in sentence 1, from 1 - 5 </rating>\newline
<thinking> Think step by step first before editing citations for sentence 2. </thinking>\newline
<editing sentence\_id="2"> edits for citations in sentence 2, or N/A if no editing is needed </editing>\newline
<rating sentence\_id="2"> rating for citations in sentence 2, from 1 - 5 </rating>\newline
...\newline
<thinking> Think step by step first before editing citations for sentence N. </thinking>\newline
<editing sentence\_id="N"> edits for citations in sentence N, or N/A if no editing is needed </editing>\newline
<rating sentence\_id="N"> rating for citations in sentence N, from 1 - 5 </rating>\newline

Begin!\newline
\end{minipage}
\end{tcolorbox}
\caption*{Table \ref{tab:prompt_citation_editing}: Continued.}
\end{table*}

%% file: acl_latex.bbl
\begin{thebibliography}{32}
\providecommand{\natexlab}[1]{#1}

\bibitem[{Bajaj et~al.(2018)Bajaj, Campos, Craswell, Deng, Gao, Liu, Majumder, McNamara, Mitra, Nguyen, Rosenberg, Song, Stoica, Tiwary, and Wang}]{msmarco}
Payal Bajaj, Daniel Campos, Nick Craswell, Li~Deng, Jianfeng Gao, Xiaodong Liu, Rangan Majumder, Andrew McNamara, Bhaskar Mitra, Tri Nguyen, Mir Rosenberg, Xia Song, Alina Stoica, Saurabh Tiwary, and Tong Wang. 2018.
\newblock \href {https://arxiv.org/abs/1611.09268} {{MS MARCO}: A human generated machine reading comprehension dataset}.
\newblock \emph{Preprint}, arXiv:1611.09268.

\bibitem[{Fan et~al.(2019)Fan, Jernite, Perez, Grangier, Weston, and Auli}]{fan-etal-2019-eli5}
Angela Fan, Yacine Jernite, Ethan Perez, David Grangier, Jason Weston, and Michael Auli. 2019.
\newblock \href {https://doi.org/10.18653/v1/P19-1346} {{ELI}5: Long form question answering}.
\newblock In \emph{Proceedings of the 57th Annual Meeting of the Association for Computational Linguistics}, pages 3558--3567, Florence, Italy.

\bibitem[{Fierro et~al.(2024)Fierro, Amplayo, Huot, De~Cao, Maynez, Narayan, and Lapata}]{fierro-etal-2024-learning}
Constanza Fierro, Reinald~Kim Amplayo, Fantine Huot, Nicola De~Cao, Joshua Maynez, Shashi Narayan, and Mirella Lapata. 2024.
\newblock \href {https://doi.org/10.18653/v1/2024.acl-long.615} {Learning to plan and generate text with citations}.
\newblock In \emph{Proceedings of the 62nd Annual Meeting of the Association for Computational Linguistics (Volume 1: Long Papers)}, pages 11397--11417, Bangkok, Thailand.

\bibitem[{Gao et~al.(2023{\natexlab{a}})Gao, Dai, Pasupat, Chen, Chaganty, Fan, Zhao, Lao, Lee, Juan, and Guu}]{gao-etal-2023-rarr}
Luyu Gao, Zhuyun Dai, Panupong Pasupat, Anthony Chen, Arun~Tejasvi Chaganty, Yicheng Fan, Vincent Zhao, Ni~Lao, Hongrae Lee, Da-Cheng Juan, and Kelvin Guu. 2023{\natexlab{a}}.
\newblock \href {https://doi.org/10.18653/v1/2023.acl-long.910} {{RARR}: Researching and revising what language models say, using language models}.
\newblock In \emph{Proceedings of the 61st Annual Meeting of the Association for Computational Linguistics (Volume 1: Long Papers)}, pages 16477--16508, Toronto, Canada.

\bibitem[{Gao et~al.(2023{\natexlab{b}})Gao, Yen, Yu, and Chen}]{gao-etal-2023-alce}
Tianyu Gao, Howard Yen, Jiatong Yu, and Danqi Chen. 2023{\natexlab{b}}.
\newblock \href {https://doi.org/10.18653/v1/2023.emnlp-main.398} {Enabling large language models to generate text with citations}.
\newblock In \emph{Proceedings of the 2023 Conference on Empirical Methods in Natural Language Processing}, pages 6465--6488, Singapore.

\bibitem[{Han et~al.(2024)Han, Zhang, Qi, Xu, Wang, Liu, Wang, Min, and Castelli}]{ragqaarena}
Rujun Han, Yuhao Zhang, Peng Qi, Yumo Xu, Jenyuan Wang, Lan Liu, William~Yang Wang, Bonan Min, and Vittorio Castelli. 2024.
\newblock \href {https://doi.org/10.18653/v1/2024.emnlp-main.249} {{RAG}-{QA} arena: Evaluating domain robustness for long-form retrieval augmented question answering}.
\newblock In \emph{Proceedings of the 2024 Conference on Empirical Methods in Natural Language Processing}, pages 4354--4374, Miami, Florida, USA.

\bibitem[{Jiang et~al.(2024)Jiang, Sablayrolles, Roux, Mensch, and et~al.}]{mixtral}
Albert~Q. Jiang, Alexandre Sablayrolles, Antoine Roux, Arthur Mensch, and et~al. 2024.
\newblock \href {https://arxiv.org/pdf/2401.04088} {Mixtral of experts}.
\newblock arXiv:2401.04088.

\bibitem[{Jiang et~al.(2023)Jiang, Xu, Gao, Sun, Liu, Dwivedi-Yu, Yang, Callan, and Neubig}]{jiang-etal-2023-active}
Zhengbao Jiang, Frank Xu, Luyu Gao, Zhiqing Sun, Qian Liu, Jane Dwivedi-Yu, Yiming Yang, Jamie Callan, and Graham Neubig. 2023.
\newblock \href {https://doi.org/10.18653/v1/2023.emnlp-main.495} {Active retrieval augmented generation}.
\newblock In \emph{Proceedings of the 2023 Conference on Empirical Methods in Natural Language Processing}, pages 7969--7992, Singapore.

\bibitem[{Karpukhin et~al.(2020)Karpukhin, Oguz, Min, Lewis, Wu, Edunov, Chen, and Yih}]{karpukhin-etal-2020-dense}
Vladimir Karpukhin, Barlas Oguz, Sewon Min, Patrick Lewis, Ledell Wu, Sergey Edunov, Danqi Chen, and Wen-tau Yih. 2020.
\newblock \href {https://doi.org/10.18653/v1/2020.emnlp-main.550} {Dense passage retrieval for open-domain question answering}.
\newblock In \emph{Proceedings of the 2020 Conference on Empirical Methods in Natural Language Processing}, pages 6769--6781, Online.

\bibitem[{Katsis et~al.(2025)Katsis, Rosenthal, Fadnis, Gunasekara, Lee, Popa, Shah, Zhu, Contractor, and Danilevsky}]{katsis2025mtrag}
Yannis Katsis, Sara Rosenthal, Kshitij Fadnis, Chulaka Gunasekara, Young-Suk Lee, Lucian Popa, Vraj Shah, Huaiyu Zhu, Danish Contractor, and Marina Danilevsky. 2025.
\newblock \href {https://arxiv.org/abs/2501.03468} {{MTRAG}: A multi-turn conversational benchmark for evaluating retrieval-augmented generation systems}.
\newblock \emph{Preprint}, arXiv:2501.03468.

\bibitem[{Lewis et~al.(2020)Lewis, Perez, Piktus, Petroni, Karpukhin, Goyal, K\"{u}ttler, Lewis, Yih, Rockt\"{a}schel, Riedel, and Kiela}]{lewis2020rag}
Patrick Lewis, Ethan Perez, Aleksandra Piktus, Fabio Petroni, Vladimir Karpukhin, Naman Goyal, Heinrich K\"{u}ttler, Mike Lewis, Wen-tau Yih, Tim Rockt\"{a}schel, Sebastian Riedel, and Douwe Kiela. 2020.
\newblock \href {https://proceedings.neurips.cc/paper/2020/file/6b493230205f780e1bc26945df7481e5-Paper.pdf} {Retrieval-augmented generation for knowledge-intensive nlp tasks}.
\newblock In \emph{Proceedings of the 34th International Conference on Neural Information Processing Systems}, Vancouver, BC, Canada.

\bibitem[{Lipson(2011)}]{lipson2011cite}
Charles Lipson. 2011.
\newblock \href {https://press.uchicago.edu/ucp/books/book/chicago/C/bo29143248.html} {\emph{Cite right: a quick guide to citation styles--MLA, APA, Chicago, the sciences, professions, and more}}.
\newblock University of Chicago Press.

\bibitem[{Liu et~al.(2023)Liu, Zhang, and Liang}]{liu-etal-2023-verifiability}
Nelson Liu, Tianyi Zhang, and Percy Liang. 2023.
\newblock \href {https://doi.org/10.18653/v1/2023.findings-emnlp.467} {Evaluating verifiability in generative search engines}.
\newblock In \emph{Findings of the Association for Computational Linguistics: EMNLP 2023}, pages 7001--7025, Singapore.

\bibitem[{Malaviya et~al.(2024)Malaviya, Lee, Chen, Sieber, Yatskar, and Roth}]{malaviya-etal-2024-expertqa}
Chaitanya Malaviya, Subin Lee, Sihao Chen, Elizabeth Sieber, Mark Yatskar, and Dan Roth. 2024.
\newblock \href {https://doi.org/10.18653/v1/2024.naacl-long.167} {{E}xpert{QA}: Expert-curated questions and attributed answers}.
\newblock In \emph{Proceedings of the 2024 Conference of the North American Chapter of the Association for Computational Linguistics: Human Language Technologies (Volume 1: Long Papers)}, pages 3025--3045, Mexico City, Mexico.

\bibitem[{Menick et~al.(2022)Menick, Trebacz, Mikulik, Aslanides, Song, Chadwick, Glaese, Young, Campbell-Gillingham, Irving, and McAleese}]{menick2022teachinglanguagemodelssupport}
Jacob Menick, Maja Trebacz, Vladimir Mikulik, John Aslanides, Francis Song, Martin Chadwick, Mia Glaese, Susannah Young, Lucy Campbell-Gillingham, Geoffrey Irving, and Nat McAleese. 2022.
\newblock \href {https://arxiv.org/abs/2203.11147} {Teaching language models to support answers with verified quotes}.
\newblock \emph{Preprint}, arXiv:2203.11147.

\bibitem[{MetaAI(2024)}]{llama-3}
MetaAI. 2024.
\newblock \href {https://arxiv.org/abs/2407.21783} {The llama 3 herd of models}.
\newblock \emph{Preprint}, arXiv:2407.21783.

\bibitem[{Muller et~al.(2023)Muller, Wieting, Clark, Kwiatkowski, Ruder, Soares, Aharoni, Herzig, and Wang}]{muller-etal-2023-evaluating}
Benjamin Muller, John Wieting, Jonathan Clark, Tom Kwiatkowski, Sebastian Ruder, Livio Soares, Roee Aharoni, Jonathan Herzig, and Xinyi Wang. 2023.
\newblock \href {https://doi.org/10.18653/v1/2023.emnlp-main.10} {Evaluating and modeling attribution for cross-lingual question answering}.
\newblock In \emph{Proceedings of the 2023 Conference on Empirical Methods in Natural Language Processing}, pages 144--157, Singapore.

\bibitem[{Rashkin et~al.(2023)Rashkin, Nikolaev, Lamm, Aroyo, Collins, Das, Petrov, Tomar, Turc, and Reitter}]{rashkin-etal-2023-ais}
Hannah Rashkin, Vitaly Nikolaev, Matthew Lamm, Lora Aroyo, Michael Collins, Dipanjan Das, Slav Petrov, Gaurav~Singh Tomar, Iulia Turc, and David Reitter. 2023.
\newblock \href {https://doi.org/10.1162/coli_a_00486} {Measuring attribution in natural language generation models}.
\newblock \emph{Computational Linguistics}, 49(4):777--840.

\bibitem[{Santhanam et~al.(2022)Santhanam, Khattab, Saad-Falcon, Potts, and Zaharia}]{santhanam-etal-2022-colbertv2}
Keshav Santhanam, Omar Khattab, Jon Saad-Falcon, Christopher Potts, and Matei Zaharia. 2022.
\newblock \href {https://doi.org/10.18653/v1/2022.naacl-main.272} {{C}ol{BERT}v2: Effective and efficient retrieval via lightweight late interaction}.
\newblock In \emph{Proceedings of the 2022 Conference of the North American Chapter of the Association for Computational Linguistics: Human Language Technologies}, pages 3715--3734, Seattle, United States.

\bibitem[{Snell et~al.(2024)Snell, Lee, Xu, and Kumar}]{snell2024scalingllmtesttimecompute}
Charlie Snell, Jaehoon Lee, Kelvin Xu, and Aviral Kumar. 2024.
\newblock \href {https://arxiv.org/abs/2408.03314} {Scaling llm test-time compute optimally can be more effective than scaling model parameters}.
\newblock \emph{Preprint}, arXiv:2408.03314.

\bibitem[{Stelmakh et~al.(2022)Stelmakh, Luan, Dhingra, and Chang}]{stelmakh-etal-2022-asqa}
Ivan Stelmakh, Yi~Luan, Bhuwan Dhingra, and Ming-Wei Chang. 2022.
\newblock \href {https://doi.org/10.18653/v1/2022.emnlp-main.566} {{ASQA}: Factoid questions meet long-form answers}.
\newblock In \emph{Proceedings of the 2022 Conference on Empirical Methods in Natural Language Processing}, pages 8273--8288, Abu Dhabi, United Arab Emirates.

\bibitem[{Team(2025)}]{qwen2025qwen25technicalreport}
Qwen Team. 2025.
\newblock \href {https://arxiv.org/abs/2412.15115} {Qwen2.5 technical report}.
\newblock \emph{Preprint}, arXiv:2412.15115.

\bibitem[{Trivedi et~al.(2023)Trivedi, Balasubramanian, Khot, and Sabharwal}]{trivedi-etal-2023-interleaving}
Harsh Trivedi, Niranjan Balasubramanian, Tushar Khot, and Ashish Sabharwal. 2023.
\newblock \href {https://doi.org/10.18653/v1/2023.acl-long.557} {Interleaving retrieval with chain-of-thought reasoning for knowledge-intensive multi-step questions}.
\newblock In \emph{Proceedings of the 61st Annual Meeting of the Association for Computational Linguistics (Volume 1: Long Papers)}, pages 10014--10037, Toronto, Canada.

\bibitem[{Williams et~al.(2018)Williams, Nangia, and Bowman}]{mnli2018williams}
Adina Williams, Nikita Nangia, and Samuel Bowman. 2018.
\newblock \href {http://aclweb.org/anthology/N18-1101} {A broad-coverage challenge corpus for sentence understanding through inference}.
\newblock In \emph{Proceedings of the 2018 Conference of the North American Chapter of the Association for Computational Linguistics: Human Language Technologies, Volume 1 (Long Papers)}, pages 1112--1122, New Orleans, Louisiana, USA.

\bibitem[{Xia et~al.(2024)Xia, Xing, Du, Yang, Feng, Xu, Yin, and Xiong}]{fofo}
Congying Xia, Chen Xing, Jiangshu Du, Xinyi Yang, Yihao Feng, Ran Xu, Wenpeng Yin, and Caiming Xiong. 2024.
\newblock \href {https://doi.org/10.18653/v1/2024.acl-long.40} {{FOFO}: A benchmark to evaluate {LLM}s' format-following capability}.
\newblock In \emph{Proceedings of the 62nd Annual Meeting of the Association for Computational Linguistics (Volume 1: Long Papers)}, pages 680--699, Bangkok, Thailand.

\bibitem[{Xu and Lapata(2022)}]{xu2022latent}
Yumo Xu and Mirella Lapata. 2022.
\newblock \href {https://doi.org/10.1162/tacl_a_00480} {Document summarization with latent queries}.
\newblock \emph{Transactions of the Association for Computational Linguistics}, 10:623--638.

\bibitem[{Yue et~al.(2023)Yue, Wang, Chen, Zhang, Su, and Sun}]{yue-etal-2023-attrscore}
Xiang Yue, Boshi Wang, Ziru Chen, Kai Zhang, Yu~Su, and Huan Sun. 2023.
\newblock \href {https://doi.org/10.18653/v1/2023.findings-emnlp.307} {Automatic evaluation of attribution by large language models}.
\newblock In \emph{Findings of the Association for Computational Linguistics: EMNLP 2023}, pages 4615--4635, Singapore.

\bibitem[{Zhang et~al.(2024{\natexlab{a}})Zhang, Xu, and Perez-Beltrachini}]{zhang-etal-2024-fine}
Huajian Zhang, Yumo Xu, and Laura Perez-Beltrachini. 2024{\natexlab{a}}.
\newblock \href {https://aclanthology.org/2024.eacl-long.102/} {Fine-grained natural language inference based faithfulness evaluation for diverse summarisation tasks}.
\newblock In \emph{Proceedings of the 18th Conference of the European Chapter of the Association for Computational Linguistics (Volume 1: Long Papers)}, pages 1701--1722, St. Julian{'}s, Malta.

\bibitem[{Zhang et~al.(2024{\natexlab{b}})Zhang, Bai, Lv, Gu, Liu, Zou, Cao, Hou, Dong, Feng, and Li}]{zhang2024longcite}
Jiajie Zhang, Yushi Bai, Xin Lv, Wanjun Gu, Danqing Liu, Minhao Zou, Shulin Cao, Lei Hou, Yuxiao Dong, Ling Feng, and Juanzi Li. 2024{\natexlab{b}}.
\newblock \href {https://arxiv.org/abs/2409.02897} {Long{C}ite: Enabling llms to generate fine-grained citations in long-context qa}.
\newblock \emph{Preprint}, arXiv:2409.02897.

\bibitem[{Zhang et~al.(2024{\natexlab{c}})Zhang, Aliannejadi, Yuan, Pei, Huang, and Kanoulas}]{zhang2024fine}
Weijia Zhang, Mohammad Aliannejadi, Yifei Yuan, Jiahuan Pei, Jia-hong Huang, and Evangelos Kanoulas. 2024{\natexlab{c}}.
\newblock \href {https://aclanthology.org/2024.inlg-main.35/} {Towards fine-grained citation evaluation in generated text: A comparative analysis of faithfulness metrics}.
\newblock In \emph{Proceedings of the 17th International Natural Language Generation Conference}, pages 427--439, Tokyo, Japan.

\bibitem[{Zhang et~al.(2024{\natexlab{d}})Zhang, Rossi, Kveton, Shao, Yang, Zamani, Dernoncourt, Barrow, Yu, Kim, Zhang, Gu, Derr, Chen, Wu, Chen, Wang, Mitra, Lipka, Ahmed, and Wang}]{zhang2024personalization}
Zhehao Zhang, Ryan~A. Rossi, Branislav Kveton, Yijia Shao, Diyi Yang, Hamed Zamani, Franck Dernoncourt, Joe Barrow, Tong Yu, Sungchul Kim, Ruiyi Zhang, Jiuxiang Gu, Tyler Derr, Hongjie Chen, Junda Wu, Xiang Chen, Zichao Wang, Subrata Mitra, Nedim Lipka, Nesreen Ahmed, and Yu~Wang. 2024{\natexlab{d}}.
\newblock \href {https://arxiv.org/abs/2411.00027} {Personalization of large language models: A survey}.
\newblock \emph{Preprint}, arXiv:2411.00027.

\bibitem[{Zheng et~al.(2023)Zheng, Chiang, Sheng, Zhuang, Wu, Zhuang, Lin, Li, Li, Xing, Zhang, Gonzalez, and Stoica}]{zheng2023judging}
Lianmin Zheng, Wei-Lin Chiang, Ying Sheng, Siyuan Zhuang, Zhanghao Wu, Yonghao Zhuang, Zi~Lin, Zhuohan Li, Dacheng Li, Eric Xing, Hao Zhang, Joseph~E. Gonzalez, and Ion Stoica. 2023.
\newblock \href {https://openreview.net/forum?id=uccHPGDlao} {Judging {LLM}-as-a-judge with {MT}-bench and chatbot arena}.
\newblock In \emph{Thirty-seventh Conference on Neural Information Processing Systems Datasets and Benchmarks Track}, New Orleans, Louisiana, USA.

\end{thebibliography}
